\pdfoutput=1
\documentclass[11pt]{article}
\usepackage[final]{acl}

\usepackage{newtxtext,newtxmath}
\usepackage{latexsym}

\usepackage[T1]{fontenc}

\usepackage[utf8]{inputenc}

\usepackage{microtype}

\usepackage{inconsolata}

\usepackage{graphicx}

\usepackage{tcolorbox}
\usepackage{booktabs}
\usepackage{colortbl}
\usepackage{multirow}
\usepackage{graphicx}
\usepackage{algorithmic,algorithm}
\usepackage{xcolor}

\definecolor{ourscolor}{gray}{0.9}

\title{CTRL: Control-Based Time Series Forecasting\\ with LLM-Guided  Residual Learning}

\author{
  \textbf{Minkyoung Kim\textsuperscript{1}},
  \textbf{Daeun Ji\textsuperscript{1}},
  \textbf{Yohan Lee\textsuperscript{1}},
  \textbf{Beomsoo Kim\textsuperscript{1}$^\dagger$},
  \textbf{Beakcheol Jang\textsuperscript{1}$^\dagger$}
\\
\\
  \textsuperscript{1}Graduate School of Information, Yonsei University, Seoul, Republic of Korea
\\
 {{\{\href{mailto:minky@yonsei.ac.kr}{\textcolor{black}{minky}}, \href{mailto:daeun09@yonsei.ac.kr}{\textcolor{black}{daeun09}}, \href{mailto:utopidamath@yonsei.ac.kr}{\textcolor{black}{utopiamath}}, \href{mailto:beomsoo@yonsei.ac.kr}{\textcolor{black}{beomsoo}}, \href{mailto:bjang@yonsei.ac.kr}{\textcolor{black}{bjang}}\}@yonsei.ac.kr}}
}

\begin{document}
\maketitle

\begingroup
\renewcommand\thefootnote{}
\footnotetext{$^\dagger$Corresponding authors.} %
\endgroup

\begin{abstract}
Time series forecasting underpins critical decision-making across diverse domains. While large language models (LLMs) offer promising reasoning capabilities, existing LLM-based time series forecasting approaches either reduce them to numerical predictors that bypass their strengths, or allow direct forecast generation that destabilizes predictions in non-stationary settings. We introduce CTRL, a framework that decouples semantic reasoning from quantitative prediction. A frozen backbone generates base forecasts, while specialized LLM agents function as controllers that analyze backbone prediction errors through decomposed trend, seasonal, and irregular components, grounding reasoning in interpretable temporal structure. Each agent outputs compact control signals that a lightweight residual decoder translates into forecast corrections. CTRL incorporates label-free test-time adaptation that detects distribution shift from input statistics alone and readapts control signals with only 3--24 LLM calls via caching. CTRL is explicitly designed to improve robustness under non-stationary temporal dynamics and distribution shift, while remaining competitive on highly stationary time series where adaptive correction provides limited additional benefit.
\end{abstract}

\section{Introduction}
Time series forecasting (TSF) supports data-driven decision-making in public health, finance, manufacturing, and energy. The emergence of large language models (LLMs) has led to growing interest in combining linguistic reasoning with temporal modeling to establish multimodal connections between numerical sequences and textual knowledge~\cite{zhou2023onefitsall,Chang2023LLM4TS,Sun2023TEST,Jin2024TimeLLM,Liu2025TimeCMA,Cao2024TEMPO,Liu2024UniTime,Pan2024S2IP,Liu2024AutoTimes,Liu2024STLLM,Liang2024Foundation}. These models leverage long-range dependencies and semantic abstraction to uncover latent temporal dynamics that traditional approaches may miss under non-stationary and event-driven environments~\cite{Liu2023AdaptiveNorm,Kim2022RevIN,Liu2023Koopa,Wen2023OneNet}.

\begin{figure}[t]
\centering
\includegraphics[width=\columnwidth]{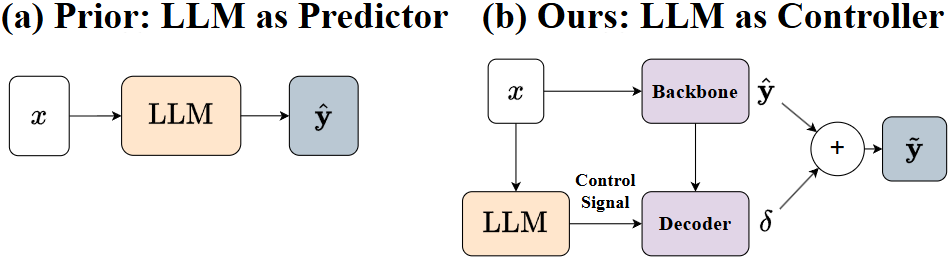}
\caption{LLM integration paradigms. (a) Prior methods use LLMs as direct numerical predictors. (b) CTRL uses LLMs as controllers generating compact control signals for a residual decoder; a frozen backbone provides base forecasts. Final prediction: $\tilde{\mathbf{y}} = \hat{\mathbf{y}} + \delta$.}
\label{fig:motivation}
\end{figure}

Existing LLM-based forecasting frameworks face two structural limitations~\cite{Zhang2025DoesMultimodality,Singh2024TokenizationCounts}.
First, many prompt-based methods reduce LLMs to Transformer-like predictors that tokenize numerical sequences and output values without exploiting their capacity for contextual reasoning and dynamic adaptation.
Second, when LLMs directly generate numerical outputs, they can disrupt the stability of time-series backbones, amplifying noise and weakening generalization in complex temporal settings.

Recent models such as TEMPO~\cite{Cao2024TEMPO} mitigate these issues by decomposing time series into trend, seasonality, and irregular components and learning them through prompt representations. Alignment-aware frameworks, including CALF~\cite{Liu2025CALF} and TimeCMA~\cite{Liu2025TimeCMA}, further improve semantic coherence through structured prompting and cross-modal alignment. %
These efforts enhance interpretability but still couple LLM representations directly to forecast generation.

This study introduces CTRL (Control-based Time Series Forecasting), a framework that separates semantic reasoning from quantitative prediction. A frozen backbone produces base forecasts, while specialized LLM agents function as controllers analyzing distinct temporal components: trend, seasonality, and irregular patterns. CTRL is motivated by the observation that many real-world time series are inherently
non-stationary, exhibiting distribution shift and regime changes where static or
alignment-based forecasting models struggle to adapt. Rather than attempting direct numerical generation, agents engage in comparative reasoning, diagnosing discrepancies between predictions and ground truth. Each agent outputs a compact control signal vector encoding scale, bias, gate, and confidence; the Irregular Agent produces natural language analysis projected via a frozen encoder. A lightweight residual decoder translates these policies into fine-grained corrections refining backbone forecasts.

CTRL performs test-time adaptation by detecting distribution shift from input statistics alone, without relying on ground-truth labels, and operates efficiently with low LLM overhead by leveraging clustered held-out training samples. By separating reasoning from prediction, the framework enables stable and extensible forecasting: LLM agents provide policy-level guidance, the decoder ensures numerical robustness, and the backbone preserves accuracy. CTRL yields the largest gains under non-stationary dynamics while remaining competitive on stationary datasets, where adaptive reasoning offers diminishing returns.

Our contributions are as follows:

\begin{itemize}
\item We position LLMs as strategic controllers that require no parameter updates, separating temporal reasoning from numerical computation. The lightweight residual decoder (${\sim}$400K parameters) is the only component trained via gradient descent.

\item Specialized agents analyze backbone errors through STL-decomposed comparisons, generating correction policies from few-shot reasoning.

\item We introduce test-time adaptation that detects distribution shift from input statistics alone, enabling LLM-guided control signal adjustment without ground truth, gradients, or parameter updates.

\item CTRL improves two architecturally distinct backbones (linear and transformer) with 3--24 LLM calls, with gains scaling with distribution shift.
\end{itemize}

\section{Related Work}

\subsection{LLM-based Forecasting}
Recent work integrates LLMs into forecasting through diverse paradigms. Reprogramming approaches convert time series into text-compatible representations processed by partially-tuned LLMs~\cite{Jin2024TimeLLM,zhou2023onefitsall}, while PromptCast~\cite{Xue2023PromptCast} formulates forecasting as natural language completion. TEMPO~\cite{Cao2024TEMPO} decomposes time series into trend, seasonal, and residual components with prompt-specific representations. Cross-modal alignment frameworks including CALF~\cite{Liu2025CALF} and TimeCMA~\cite{Liu2025TimeCMA} bridge semantic embeddings with temporal features through fine-tuning.

Most approaches position LLMs as numerical predictors, either through tokenized sequence completion or embedding alignment, limiting exploitation of their comparative reasoning capabilities. CTRL instead leverages LLMs for diagnosing discrepancies and generating strategic guidance, while delegating numerical computation to specialized decoders. Studies integrating exogenous textual information~\cite{Zhou2025Unveiling,Jiang2025Explainable} depend on fixed representations that fail to align with evolving temporal dynamics. Recent empirical work~\cite{Tan2024LLMUseful,park2025revisiting} questions the value of LLMs as direct numerical forecasters. Alignment-based methods assume stable temporal distributions, whereas CTRL explicitly targets non-stationary settings through test-time adaptation.

\subsection{Training-Free LLM Approaches}

In-context learning enables frozen LLMs to adapt without gradient updates~\cite{brown2020gpt3}, while chain-of-thought prompting~\cite{wei2022cot} activates multi-step reasoning. Recent work frames LLMs as reasoning agents invoking external modules for precise computation~\cite{schick2023toolformer,yao2023react}, separating planning from execution.
In forecasting, FPT~\cite{zhou2023onefitsall} demonstrates that GPT-2 with frozen self-attention layers performs competitively. LLMTime~\cite{gruver2023llmtime} frames forecasting as next-token prediction, enabling zero-shot forecasting competitive with supervised baselines. However, these approaches utilize LLMs primarily for pattern mapping without exploiting reasoning capabilities for temporal dynamics.

\textsc{CTRL} extends this paradigm by activating reasoning through structured STL-decomposed context. The LLM serves as a diagnostic reasoner generating interpretable control signals, while numerical computation is delegated to a residual decoder, enabling adaptation to distribution shift through explicit temporal analysis.

Prior approaches to distribution shift in forecasting typically require additional mechanisms. Post-hoc methods such as ResCAL~\cite{kim2022rescal} exploit autocorrelated prediction errors but train a separate estimation module. Gradient-based test-time adaptation including TENT~\cite{wang2021tent} and SAR~\cite{niu2023sar} updates model parameters during inference via entropy minimization, constituting a different evaluation protocol from standard forecasting benchmarks. CTRL instead performs label-free adaptation without gradient computation or parameter updates, detecting shift from input statistics and adjusting policies through LLM reasoning.

\begin{figure*}[t]
\centering
\includegraphics[width=1\textwidth]{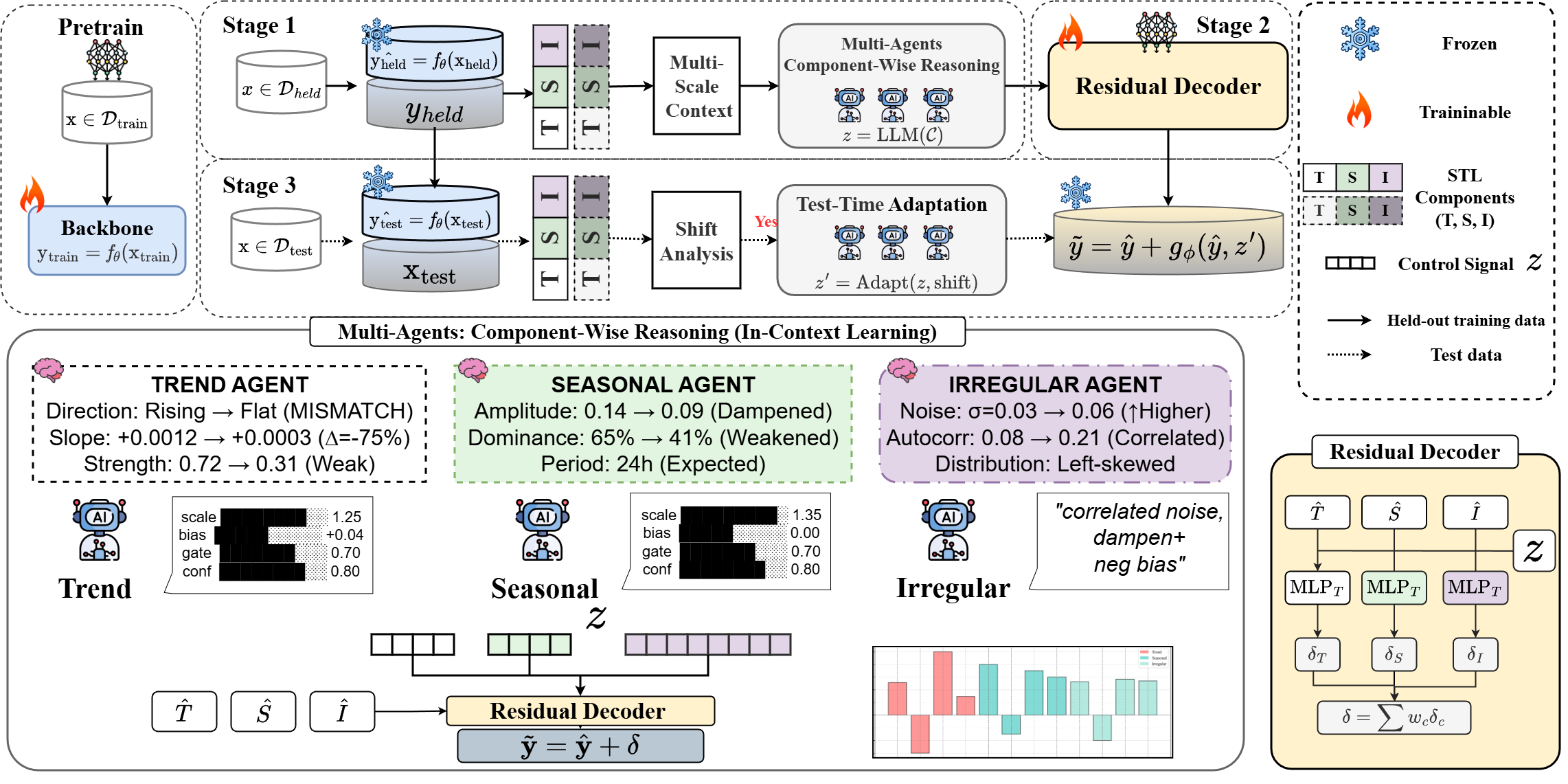}
\caption{\textbf{CTRL Framework Overview.} A frozen backbone generates initial predictions $\hat{y}$. In Stage 1, three specialized LLM agents analyze STL-decomposed held-out training statistics via in-context reasoning: Trend and Seasonal agents output 4D control signals [scale, bias, gate, confidence], while the Irregular agent produces text encoded via frozen GPT-2. In Stage 2, concatenated control signals guide component-specific MLPs in the residual decoder, trained on held-out training data with ground truth. In Stage 3, test-time adaptation detects distribution shift from input statistics alone and adjusts control signals without accessing ground-truth labels. Final prediction: $\tilde{y} = \hat{y} + \delta$.}
\label{fig:pipeline}
\end{figure*}

\section{Methodology}
\paragraph{Notation.} We denote input sequences as $\mathbf{x} \in \mathbb{R}^{L \times F}$ with lookback $L$ and $F$ features, targets as $\mathbf{y} \in \mathbb{R}^{H \times F}$ with horizon $H$. Backbone predictions are $\hat{\mathbf{y}}$, refined predictions $\tilde{\mathbf{y}}$. Superscripts $(T), (S), (I)$ index trend, seasonal, and irregular components respectively.

\subsection{Overview}

\begin{algorithm}[h!]
\small
\caption{CTRL: LLM-Guided Residual Learning}
\label{alg:ctrl}
\begin{algorithmic}[1]
\REQUIRE Frozen backbone $f_\theta$, LLM agents, threshold $\tau{=}2.0$, check interval $K{=}50$
\ENSURE Refined prediction $\tilde{y} = \hat{y} + \delta$

\STATE \textcolor{blue}{\textit{// Stage 1: Control Signal Generation (no LLM parameter updates)}}
\STATE Compute global context: dataset-level error statistics
\STATE Cluster held-out training samples by temporal features via TypiClust
\STATE Retrieve $k{=}16$ few-shot examples with measured backbone errors
\STATE Compute current context: STL comparison of prediction vs ground truth
\FOR{agent $\in$ \{Trend, Seasonal\}}
    \STATE $z^{(\cdot)} \gets$ [scale, bias, gate, conf] via in-context reasoning
\ENDFOR
\STATE $z^{(I)} \gets$ encode Irregular agent text via frozen GPT-2
\STATE All samples share the same control signal $z$ \COMMENT{3 LLM calls total}

\STATE \textcolor{blue}{\textit{// Stage 2: Residual Decoder Training}}
\FOR{each $(x, y)$ in held-out training set}
    \STATE $\hat{y} \gets f_\theta(x)$
    \STATE $(\hat{T}, \hat{S}, \hat{I}) \gets \textsc{STL}(\hat{y})$
    \STATE $\delta^{(c)} \gets \text{MLP}_c([\hat{c}, z])$ for $c \in \{T, S, I\}$
    \STATE $\delta \gets w_T \delta^{(T)} + w_S \delta^{(S)} + w_I \delta^{(I)}$
    \STATE Minimize $\|\hat{y} + \delta - y\|^2$
\ENDFOR

\STATE \textcolor{blue}{\textit{// Stage 3: Test-Time Adaptation (every $K$ batches; Irregular bypasses)}}
\STATE Cache STL statistics from held-out training inputs
\FOR{each test batch}
    \STATE $\hat{y} \gets f_\theta(x_{\text{test}})$
    \STATE $s_c \gets$ z-score of $\textsc{STL}(x_{\text{test}})$ vs held-out training
    \IF{$s_c > \tau$ for $c \in \{T, S\}$}
        \STATE $z^{(c)} \gets$ LLM-adjusted control signal
    \ENDIF
    \STATE $\tilde{y} \gets \hat{y} + g_\phi(\hat{y}, z)$
\ENDFOR
\end{algorithmic}
\end{algorithm}

CTRL refines backbone forecasts through LLM-guided residual learning. LLMs excel at grounded reasoning, diagnosing discrepancies between predictions and ground truth, but struggle with precise numerical prediction. We exploit this asymmetry: LLM agents function as controllers that analyze prediction errors and generate compact policies guiding a learned residual decoder, while the backbone maintains numerical precision.

Given a multivariate time series with input $\mathbf{x} \in \mathbb{R}^{L \times F}$ and target $\mathbf{y} \in \mathbb{R}^{H \times F}$, a pretrained backbone $f_\theta$ produces initial predictions $\hat{\mathbf{y}} = f_\theta(\mathbf{x})$. CTRL refines these via a control signal-guided residual decoder:
\begin{equation}
    \tilde{\mathbf{y}} = \hat{\mathbf{y}} + g_\phi(\hat{\mathbf{y}}, \mathbf{z})
\end{equation}
where $g_\phi$ is the residual decoder with parameters $\phi$, and $\mathbf{z} \in \mathbb{R}^{d_z}$ is the control signal vector generated by LLM agents ($d_z = 8 + d_I$, where $d_I$ is the text projection dimension with default $d_I = 32$).

As shown in Figure \ref{fig:pipeline} and Algorithm~\ref{alg:ctrl}, CTRL operates in three stages: (1) control signal generation, where LLM agents produce correction policies via in-context reasoning without any LLM parameter updates; (2) residual decoder training on held-out training data; and (3) test-time adaptation that adjusts policies when distribution shift is detected.
Critically, test-time adaptation requires \emph{no ground-truth labels}, enabling adaptation to distribution shifts using only input statistics. LLM agents require no fine-tuning. Policies emerge from few-shot in-context reasoning over STL statistics. %
\subsection{Backbone Pretraining}

We train a time series backbone on training data, then freeze its weights. We evaluate DLinear~\citep{Zeng2023DLinear} (linear) and PatchTST~\citep{Nie2023PatchTST} (transformer), representing two dominant architectural paradigms in time series forecasting.
\subsection{Multi-Agent Control Signal Generation}

Control signal generation analyzes \emph{how} the backbone fails by comparing predictions against ground truth on held-out training data. 
Rather than using LLMs as numerical predictors, we position them as controllers that generate strategic correction policies while delegating computation to the residual decoder. Each agent receives multi-scale context to ground its reasoning.
This process requires no LLM parameter updates: agents perform in-context control signal generation by reasoning over held-out training statistics in a single forward pass. No gradients flow through the LLM at any stage; only the residual decoder is trained.

Effective control signal generation requires grounding at multiple scales. At the global level, agents receive dataset-wide statistics, such as mean error magnitudes, trend directions, seasonal strength distributions, establishing baseline expectations. Local context provides few-shot examples from similar held-out training samples, showing concrete failure patterns with measured errors that adapt correction magnitudes. 
Current context supplies STL decomposition comparing the sample's prediction against ground truth, including trend slope, seasonal amplitude, and dominance metrics (Appendix~\ref{app:stl_context}).

We select examples via TypiClust~\cite{hacohen2022active}, clustering held-out training samples by temporal features $\mathbf{f}_i = [\text{slope}, \sigma, \rho_1, S_{\text{strength}}]$ (trend slope, volatility, first-order autocorrelation, seasonal strength). We retrieve $k=16$ examples prioritizing same-cluster members while including diverse contrasts.

These held-out training examples include ground truth, enabling display of \emph{actual} backbone errors:
\begin{quote}
\small
\textit{``Similar pattern (upward trend, high volatility): Backbone MAE was 0.082, underestimating trend slope by 15\%.''}
\end{quote}
This grounds control signal decisions in empirical magnitudes rather than pattern inference alone.

We decompose both backbone prediction and ground truth into trend, seasonal, and irregular components using STL, providing agents with structured comparison of temporal characteristics.
STL decomposition partitions prediction errors into distinct frequency bands, each exhibiting characteristic failure modes that benefit from specialized analysis. The {Trend Agent} examines low-frequency dynamics including direction, level shifts, and systematic bias, detecting momentum underestimation when backbones predict flat trends against rising ground truth. The {Seasonal Agent} targets periodic patterns, diagnosing amplitude dampening and phase misalignment that backbones often introduce through oversmoothing. The {Irregular Agent} assesses high-frequency noise characteristics, prescribing adjustments when backbones either over-smooth or amplify fluctuations, enabling each agent to develop focused expertise. %

Each numerical agent (Trend, Seasonal) outputs a 4-dimensional control signal vector:
\begin{equation}
    \mathbf{z}^{(c)} = [\texttt{scale}, \texttt{bias}, \texttt{gate}, \texttt{confidence}] \in \mathbb{R}^4
\end{equation}
where $c \in \{T, S\}$ indexes trend and seasonal components, and \texttt{scale} $\in [0.5, 1.5]$ and \texttt{bias} $\in [-1, 1]$ provide linear modulation~\cite{perez2018film}, \texttt{gate} $\in [0, 1]$ controls correction aggressiveness inspired by gating mechanisms~\cite{hochreiter1997lstm}, and \texttt{confidence} $\in [0, 1]$ expresses agent certainty for uncertainty-aware weighting~\cite{xiong2024llms-conf,geng2024survey-conf}.

The Irregular Agent produces natural language analysis encoding uncertainty and anomaly patterns, projected to a $d_I$-dimensional embedding via a frozen text encoder:
\begin{equation}
    \mathbf{z}^{(I)} = \texttt{TextProj}(\texttt{GPT-2}(\mathbf{t}^{(I)})) \in \mathbb{R}^{d_I}
\end{equation}
where \texttt{TextProj} is a lightweight MLP projector that maps the 768D GPT-2 embedding to $d_I$ dimensions.

Prior work produces high-dimensional embeddings that downstream projections largely discard~\cite{Tan2024LLMUseful}. Our bounded output aligns with LLM strengths, structured values with clear semantics, while forcing distillation of actionable guidance.

\paragraph{Structural isolation from pretraining data.} LLM agents never receive raw time series values or dataset identifiers. Their inputs consist solely of backbone-specific runtime statistics: per-sample error magnitudes, slope deviations between ground truth and a specific backbone's predictions on a specific held-out split, seasonal amplitude mismatches, and dominance ratios. Concrete examples include phrases such as ``DLinear underestimates trend slope by 15\%'' or ``seasonal amplitude dampened from 0.14 to 0.09'', which are generated at runtime and depend on the backbone instance, the held-out split, and the STL decomposition parameters. These artifacts cannot exist in any pretraining corpus, making data leakage through LLM memorization architecturally precluded. This is supported empirically: removing STL context from prompts degrades performance by +2.11\%/+3.10\% (Table~\ref{tab:ablation}), confirming the LLM reasons over provided inputs rather than retrieving memorized patterns.

\begin{table*}[h!]
\centering
\footnotesize
\setlength{\tabcolsep}{1.1pt}
\newcolumntype{b}{>{\columncolor{cyan!5}}c}
\begin{tabular}{clbbbbcccccccccccc}
\toprule
\textbf{Data} & \textbf{H} & \multicolumn{2}{c}{\cellcolor{cyan!5}\textbf{Ours(DL)}} & \multicolumn{2}{c}{\cellcolor{cyan!5}\textbf{Ours(PT)}} & \multicolumn{2}{c}{\textbf{DLinear}} & \multicolumn{2}{c}{\textbf{PatchTST}} & \multicolumn{2}{c}{\textbf{TimeLLM$^\dagger$}} & \multicolumn{2}{c}{\textbf{CALF}} & \multicolumn{2}{c}{\textbf{TEMPO}} & \multicolumn{2}{c}{\textbf{GPT4TS}} \\
& & \cellcolor{cyan!5}\textbf{MAE} & \cellcolor{cyan!5}\textbf{MSE} & \cellcolor{cyan!5}\textbf{MAE} & \cellcolor{cyan!5}\textbf{MSE} & \textbf{MAE} & \textbf{MSE} & \textbf{MAE} & \textbf{MSE} & \textbf{MAE} & \textbf{MSE} & \textbf{MAE} & \textbf{MSE} & \textbf{MAE} & \textbf{MSE} & \textbf{MAE} & \textbf{MSE} \\
\midrule
\multirow{5}{*}{\rotatebox{90}{ETTh1}} 
& 48 & \textbf{.378} & \textbf{.340} & .381 & .342 & .382 & .346 & .386 & .346 & .398 & .369 & .400 & .372 & .419 & .395 & .425 & .412 \\
& 96 & \textbf{.398} & \textbf{.371} & .399 & .372 & .399 & .372 & .405 & .380 & .429 & .414 & .410 & .393 & .469 & .471 & .464 & .473 \\
& 192 & \textbf{.429} & \textbf{.410} & .439 & .429 & .430 & .412 & .439 & .431 & .432 & .425 & .436 & .430 & .494 & .514 & .527 & .598 \\
& 336 & .457 & .442 & \textbf{.430} & \textbf{.415} & .459 & .444 & .431 & .416 & .452 & .450 & .476 & .490 & .477 & .485 & .599 & .738 \\
& 720 & .510 & .498 & \textbf{.458} & \textbf{.443} & .511 & .497 & .459 & .444 & .477 & .462 & .522 & .540 & .502 & .502 & .600 & .783 \\
\midrule
\multirow{5}{*}{\rotatebox{90}{ETTh2}} 
& 48 & \textbf{.296} & \textbf{.217} & .309 & .242 & .318 & .235 & .309 & .242 & .318 & .243 & .316 & .243 & .334 & .263 & .343 & .280 \\
& 96 & \textbf{.335} & \textbf{.273} & .356 & .305 & .370 & .307 & .410 & .343 & .359 & .307 & .367 & .320 & .379 & .332 & .344 & .284 \\
& 192 & \textbf{.388} & \textbf{.349} & .408 & .373 & .427 & .395 & .427 & .376 & .392 & .357 & .421 & .409 & .406 & .378 & .421 & .412 \\
& 336 & .427 & .383 & \textbf{.406} & \textbf{.355} & .455 & .438 & .414 & .364 & .412 & .375 & .421 & .380 & .422 & .399 & .442 & .429 \\
& 720 & .533 & .501 & .465 & \textbf{.408} & .575 & .650 & .482 & .458 & .463 & .442 & \textbf{.442} & .416 & .462 & .436 & .477 & .475 \\
\midrule
\multirow{5}{*}{\rotatebox{90}{ETTm1}} 
& 48 & .329 & .272 & \textbf{.319} & .264 & .332 & .276 & \textbf{.319} & .268 & .324 & \textbf{.261} & .325 & .266 & .358 & .302 & .394 & .383 \\
& 96 & \textbf{.344} & \textbf{.297} & .382 & .362 & .347 & .301 & .393 & .404 & .350 & \textbf{.297} & .347 & .298 & .382 & .337 & .395 & .384 \\
& 192 & \textbf{.366} & .330 & .427 & .439 & .368 & .338 & .436 & .473 & .372 & .333 & .374 & \textbf{.320} & .408 & .390 & .424 & .423 \\
& 336 & \textbf{.388} & \textbf{.370} & .414 & .409 & .390 & .371 & .419 & .417 & .398 & .379 & .394 & .371 & .434 & .421 & .444 & .458 \\
& 720 & \textbf{.421} & .421 & .446 & .446 & .429 & .430 & .454 & .460 & .425 & \textbf{.415} & .427 & .428 & .462 & .436 & .479 & .525 \\
\midrule
\multirow{5}{*}{\rotatebox{90}{ETTm2}} 
& 48 & .226 & \textbf{.126} & \textbf{.222} & .129 & .230 & .129 & .223 & .130 & .236 & .136 & .226 & .128 & .240 & .146 & .245 & .151 \\
& 96 & \textbf{.251} & \textbf{.161} & .278 & .196 & .271 & .173 & .278 & .197 & .267 & .177 & .258 & .170 & .272 & .190 & .266 & .187 \\
& 192 & \textbf{.292} & \textbf{.214} & .315 & .247 & .311 & .230 & .319 & .256 & .309 & .238 & .296 & .226 & .320 & .271 & .315 & .260 \\
& 336 & \textbf{.328} & \textbf{.265} & .363 & .331 & .350 & .287 & .374 & .367 & .341 & .290 & .330 & .278 & .357 & .324 & .350 & .317 \\
& 720 & .386 & \textbf{.354} & .418 & .422 & .412 & .386 & .436 & .478 & .388 & .368 & \textbf{.382} & .362 & .413 & .409 & .406 & .408 \\
\midrule
\multirow{5}{*}{\rotatebox{90}{Weather}} 
& 48 & .190 & .135 & \textbf{.159} & \textbf{.118} & .196 & .137 & \textbf{.159} & .119 & .178 & .127 & .160 & .123 & .199 & .146 & .176 & .129 \\
& 96 & .225 & .169 & \textbf{.201} & \textbf{.156} & .232 & .172 & .209 & .158 & .212 & .161 & \textbf{.201} & .158 & .210 & \textbf{.156} & .217 & .167 \\
& 192 & .256 & \textbf{.200} & .265 & .232 & .276 & .216 & .275 & .243 & .250 & .202 & \textbf{.241} & \textbf{.200} & .259 & .206 & .259 & .216 \\
& 336 & .297 & .258 & .311 & .278 & .310 & .261 & .315 & .304 & .289 & .252 & \textbf{.282} & \textbf{.250} & .351 & .394 & .297 & .266 \\
& 720 & .343 & \textbf{.313} & .360 & .330 & .358 & .321 & .365 & .373 & .338 & .323 & \textbf{.330} & .320 & .347 & .339 & .343 & .330 \\
\midrule
\multirow{5}{*}{\rotatebox{90}{ECL}} 
& 48 & .226 & .127 & .205 & \textbf{.108} & .233 & .130 & .205 & .111 & .218 & .116 & \textbf{.200} & .109 & .212 & .114 & .234 & .131 \\
& 96 & .241 & .142 & \textbf{.224} & \textbf{.131} & .247 & .145 & \textbf{.224} & \textbf{.131} & .231 & .132 & .227 & .136 & .233 & .134 & .247 & .147 \\
& 192 & .254 & .155 & \textbf{.242} & \textbf{.149} & .259 & .158 & .243 & .150 & .250 & .150 & .245 & .156 & .246 & .152 & .258 & .162 \\
& 336 & .270 & .170 & \textbf{.258} & \textbf{.165} & .272 & .171 & .259 & \textbf{.165} & .266 & .167 & .263 & .171 & .264 & .171 & .275 & .179 \\
& 720 & .303 & .204 & .292 & .201 & .303 & .204 & .292 & .202 & .296 & .202 & \textbf{.276} & \textbf{.190} & .294 & .204 & .316 & .232 \\
\midrule
\multirow{5}{*}{\rotatebox{90}{Exch.}} 
& 24 & .104 & \textbf{.024} & \textbf{.101} & \textbf{.024} & .105 & \textbf{.024} & .105 & .025 & .135 & .036 & .106 & .025 & .109 & .026 & .118 & .029 \\
& 48 & .147 & \textbf{.044} & \textbf{.146} & \textbf{.044} & .150 & .045 & .149 & .048 & .164 & .053 & \textbf{.146} & \textbf{.044} & .154 & .049 & .157 & .050 \\
& 96 & \textbf{.201} & \textbf{.080} & \textbf{.201} & .083 & .208 & .082 & .212 & .094 & .234 & .107 & .206 & .087 & .207 & .088 & .249 & .121 \\
& 192 & .305 & \textbf{.168} & .308 & .172 & .325 & .186 & .311 & .196 & .349 & .229 & \textbf{.302} & .181 & .310 & .192 & .324 & .203 \\
& 336 & .418 & \textbf{.311} & \textbf{.416} & .334 & .422 & .317 & .418 & .334 & .476 & .409 & .429 & .353 & .443 & .368 & .454 & .383 \\
\bottomrule
\end{tabular}
\caption{Forecasting results on multivariate benchmarks. \textbf{Bold}: best. (DL) and (PT) indicate DLinear and PatchTST backbones, respectively. $^\dagger$Time-LLM~\cite{Jin2024TimeLLM} uses GPT-2 backend with a single seed; all other results averaged over 3 seeds.}
\label{tab:main_results}
\end{table*}

\subsection{Residual Decoder Training}

With frozen backbone and frozen LLM, we train a lightweight residual decoder on held-out training data. The decoder (${\sim}$400K parameters) is the only component updated via gradient descent, making it orders of magnitude smaller than the frozen modules it conditions on.

The decoder maintains component separation. We  apply STL decomposition to backbone predictions into trend, seasonal, and irregular components:
\begin{equation}
    \hat{\mathbf{y}} = \hat{\mathbf{T}} + \hat{\mathbf{S}} + \hat{\mathbf{I}}
\end{equation}
Three parallel MLPs process each component, receiving the concatenated control signal:
\begin{align}
    \boldsymbol{\delta}^{(T)} &= \texttt{MLP}_T([\hat{\mathbf{T}}, \mathbf{z}]) \\
    \boldsymbol{\delta}^{(S)} &= \texttt{MLP}_S([\hat{\mathbf{S}}, \mathbf{z}]) \\
    \boldsymbol{\delta}^{(I)} &= \texttt{MLP}_I([\hat{\mathbf{I}}, \mathbf{z}])
\end{align}
This enforces architectural alignment between specialized agents and their target components.

While control signal parameters are concatenated to MLP inputs rather than architecturally enforced, end-to-end training encourages the decoder to learn appropriate mappings. 

The residual decoder minimizes MSE between corrected predictions and ground truth:
\begin{equation}
    \mathcal{L} = \frac{1}{N}\sum_{i=1}^{N} \lVert \hat{\mathbf{y}}_i + \boldsymbol{\delta}_i - \mathbf{y}_i \rVert^2
\end{equation}
where the combined correction $\boldsymbol{\delta} = w_T \boldsymbol{\delta}^{(T)} + w_S \boldsymbol{\delta}^{(S)} + w_I \boldsymbol{\delta}^{(I)}$ aggregates component-wise outputs with learned weights $w_T, w_S, w_I$.

\subsection{Test-Time Adaptation}

Policies trained on held-out training must generalize to potentially shifted test distributions without ground truth labels. We address this via label-free adaptation that detects distribution shift from input statistics alone and adjusts control signal parameters accordingly.

During control signal generation, we cache STL statistics from held-out training inputs: trend slope distribution, seasonal amplitude range, and irregular variance. At test time, we compute per-component divergence between test and held-out training statistics:
\begin{align}
    s_T &= \frac{|\text{slope}_\text{test} - \bar{\text{slope}}_\text{val}|}{\sigma_{\text{slope,val}} + \epsilon} \\
    s_S &= \frac{|d_{S,\text{test}} - \bar{d}_{S,\text{val}}|}{\sigma_{d_S,\text{val}} + \epsilon}
\end{align}
where $\bar{\cdot}$ and $\sigma$ denote held-out training mean and standard deviation respectively. This z-score formulation yields an adaptive threshold: we trigger adaptation when $s_c > \tau$ with $\tau = 2.0$, corresponding to test statistics outside two standard deviations of the held-out training distribution.

When adaptation triggers for component $c \in \{T, S\}$, the LLM agent receives shift statistics and outputs adjustment factors that modulate the original control signal through the same four-parameter interface used in control signal generation:
\begin{align}
    \texttt{scale}' &= \texttt{scale} \times m_\text{scale} \\
    \texttt{bias}' &= \texttt{bias} + \Delta_\text{bias} \\
    \texttt{gate}' &= \texttt{gate} \times m_\text{gate} \\
    \texttt{conf}' &= \texttt{conf} \times (1 - p_\text{conf})
\end{align}
where multipliers $m_* \in [0.5, 1.5]$ scale proportionally to shift severity, and confidence penalty $p_\text{conf} \in [0, 0.5]$ reflects increased uncertainty under distribution shift. Unshifted components retain original policies. The irregular agent uses text embedding. At inference, the adapted control signal $\mathbf{z}'$ passes through the same residual decoder interface, ensuring architectural consistency between training and test-time adaptation.

CTRL requires 3 LLM calls for initial control signal generation (one per agent), 
with test-time adaptation adding at most 3 calls per calibration check. This yields 3--24 total calls across datasets.

\section{Experiments}
\label{sec:experiments}

\subsection{Experimental Setup}

We evaluate on standard multivariate benchmarks summarized in Table~\ref{tab:datasets}: ETT variants (ETTh1, ETTh2, ETTm1, ETTm2), Electricity (ECL), Weather, and Exchange. We use prediction horizons $H \in \{48, 96, 192, 336, 720\}$ with input length 384, and $H \in \{24, 48, 96, 192, 336\}$ for the daily frequency Exchange dataset. We evaluate CTRL with two backbone architectures: DLinear~\cite{Zeng2023DLinear} and PatchTST~\cite{Nie2023PatchTST}. The residual decoder is a 2-layer MLP with STL-augmented features, and we use Llama 3.3 70B for control signal generation. All results are averaged over 3 seeds. Full hyperparameter settings are provided in Table~\ref{tab:hyperparams}.

\subsection{Main Results}
\label{sec:main_results}
Table~\ref{tab:main_results} compares CTRL against backbone baselines and LLM-based methods across seven multivariate benchmarks. CTRL achieves best or competitive performance on most dataset-horizon combinations, with improvement magnitude correlating with train-test distribution gap.

ETTh2 and ETTm2 exhibit the strongest improvements, with CTRL achieving up to 12\% MSE reduction. Both datasets show substantial train-test divergence, where LLM-guided control signals effectively identify correction strategies for non-stationary dynamics.
On Exchange, CTRL improves over the backbones across all horizons, though gains vary with horizon length due to predictable trend patterns. For high-dimensional ECL, CTRL with PatchTST achieves best results at horizons 96--336, demonstrating that shared control signal mode scales effectively. On ETTh1 and ETTm1, CTRL improves backbones by up to 2.5\%, consistent with their lower distribution shift.
On Weather, the most stationary dataset, CALF outperforms CTRL at mid-range horizons. This is expected: when distributional shift is minimal, adaptive reasoning provides limited benefit over fine-tuned alignment.

\begin{table}[h!]
\centering
\small
\setlength{\tabcolsep}{2.4pt}
\begin{tabular}{lccccccc}
\toprule
& Exch. & \multicolumn{4}{c}{ETT} & ECL & Weather \\
\cmidrule(lr){3-6}
&      & h2   & m2   & h1   & m1   &     &     \\
\midrule
ADF   & -1.9  & -4.1 & -5.7 & -5.9 & -15.0 & -8.5 & -26.7 \\
Shift & 2.33  & 1.48 & 1.49 & 0.50 & 0.50  & 0.62 & 0.71 \\
\bottomrule
\end{tabular}
\caption{Dataset stationarity. Lower ADF indicates stronger stationarity; Shift measures train-test distribution divergence via normalized mean difference.}
\label{tab:stationarity}
\end{table}

Table~\ref{tab:stationarity} confirms that CTRL's improvements correlate with distribution shift severity. CTRL consistently outperforms TEMPO while requiring no LLM fine-tuning. 

\begin{figure}[h!]
    \centering
    \includegraphics[width=1\linewidth]{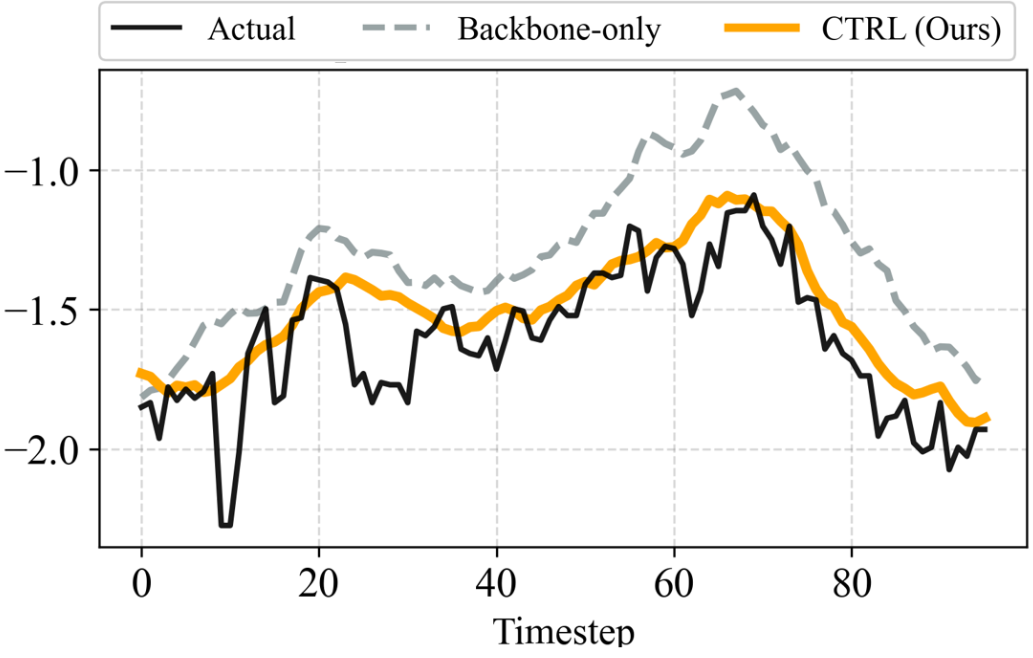}
    \caption{Backbone prediction vs CTRL-refined prediction on ETTm2}
    \label{fig:pred}
\end{figure}

Figure~\ref{fig:pred} shows CTRL correcting backbone errors during 
a distribution shift (timestep 60--80), where the backbone undershoots 
while CTRL adapts via LLM-guided control signals.

\paragraph{Comparison with direct LLM prediction.}
Table~\ref{tab:llmtime} compares CTRL against LLMTime~\cite{gruver2023llmtime}, run with GPT-4 as a direct numerical predictor through per-sample prompt-only generation. CTRL outperforms LLMTime by 19.6--44.3\% MSE, validating our core hypothesis that LLMs excel as strategic controllers rather than numerical predictors. While LLMTime requires $N$ LLM calls for $N$ test samples, CTRL achieves superior accuracy with only 3--24 calls through control signal caching.

\begin{table}[h!]
\centering
\footnotesize
\setlength{\tabcolsep}{3.9pt}
\begin{tabular}{lcccc}
\toprule
 & \multicolumn{2}{c}{\textbf{ETTm2}} & \multicolumn{2}{c}{\textbf{Weather}} \\
\cmidrule(lr){2-3} \cmidrule(lr){4-5}
 & MSE & MAE & MSE & MAE \\
\midrule
LLMTime & .245 & .151 & .237 & .150 \\
\textbf{CTRL} & \textbf{.197} & \textbf{.070} & \textbf{.132} & \textbf{.039} \\
\midrule
\textit{Improv.} & \textit{19.6\%} & \textit{53.2\%} & \textit{44.3\%} & \textit{74.2\%} \\
\bottomrule
\end{tabular}
\caption{Comparison with LLMTime. CTRL uses DLinear backbone.}
\label{tab:llmtime}
\end{table}

\subsection{Ablation Studies}
\label{sec:ablation}

\begin{table}[h!]
\centering
\resizebox{\columnwidth}{!}{
\begin{tabular}{lcc}
\toprule
\textbf{Configuration} & ETTh2 & ETTm2  \\
\midrule
\multicolumn{3}{l}{\textbf{Baseline}} \\
Backbone only & +9.45 & +7.20 \\
Learned control signal & +3.13 & +4.42 \\
Random control signal  & +5.84 & +4.54 \\   
Zero control signal & +5.70 & +5.18 \\
\midrule
\multicolumn{3}{l}{\textbf{Test-time Adaptation}} \\
w/o test-time adaptation & +1.03 & +1.79 \\
\midrule
\multicolumn{3}{l}{\textbf{Agent Ablation}} \\
Single Agent & +1.64 & +1.72 \\
w/o Trend Agent & +1.36 & +1.60 \\
w/o Seasonal Agent & +4.48 & +1.55 \\
w/o Irregular Agent & +3.37 & +1.01 \\
Irregular Numerical & +2.04 & +2.45 \\
\midrule
\multicolumn{3}{l}{\textbf{Prompt Design}} \\
w/o STL context & +2.11 & +3.10 \\
\midrule
\multicolumn{3}{l}{\textbf{Few-shot Selection}} \\
None & +1.12 & +1.59 \\
Random $k$=16 & +1.20 & +1.10 \\
\midrule
\multicolumn{3}{l}{\textbf{LLM Choice}} \\
GPT-4o & -0.34 & -0.21 \\
Qwen 2.5 72B & +1.89 & +1.63 \\
\midrule
\multicolumn{3}{l}{\textbf{Control Signal Parameters}} \\
w/o scale & +0.49 & +0.50 \\
w/o bias & +0.48 & +0.49 \\
w/o gate & +0.49 & +0.50 \\
w/o confidence & +0.50 & +0.52 \\
\bottomrule
\end{tabular}%
}
\caption{Design ablation on ETTh2 and ETTm2 (384$\rightarrow$96). $\Delta$ indicates $\Delta${MSE(\%)}, performance degradation vs. CTRL (full). Results are averaged over 3 seeds.}
\label{tab:ablation}
\end{table}

Table~\ref{tab:ablation} presents comprehensive ablations on ETTh2 and ETTm2 (384$\rightarrow$96) to validate each design choice.
The learned baseline directly maps STL features to control signals via MLP, yet underperforms LLM-generated signals, confirming that semantic reasoning provides value beyond pattern matching. Critically, we construct two random baselines: numeric vectors sampled uniformly within operational ranges, and random text fed through the same GPT-2 pipeline used by the Irregular Agent. Both degrade to the zero-signal level (+5.84\%/+4.54\% vs +5.70\%/+5.18\%), directly refuting the hypothesis that embeddings serve as generic learnable bias terms. This rules out the alternative explanation that CTRL's gains stem from increased model capacity.

\paragraph{Semantic content drives correction.} To verify the decoder exploits semantic content rather than surface features, we measure how decoder output changes when each control signal component is perturbed across its operational range (full methodology in Appendix~\ref{app:sensitivity}). The irregular text embedding yields $\Delta = 0.032$, while the most responsive numeric parameter (trend bias) yields $\Delta = 0.009$, a $3.4\times$ gap. Substituting the LLM-generated text with random words, technical jargon, or random characters produces $\Delta \leq 0.007$, an order of magnitude below the ``no text'' condition ($\Delta = 0.032$). The decoder thus distinguishes presence from absence of semantic content and responds selectively to LLM-generated descriptions, not to arbitrary text conditioning.

STL-decomposed multi-agent reasoning proves essential. Removing STL context consistently degrades performance, confirming that structured temporal decomposition enables more precise control signal generation. Agent ablations reveal dataset-dependent contributions: on ETTh2, the Seasonal Agent contributes most significantly, while ETTm2 shows more balanced contributions across agents. Converting the Irregular Agent from text-based to numerical output degrades performance, validating natural language for capturing nuanced noise characteristics.

For few-shot selection, TypiClust outperforms both random sampling and no few-shot baselines. By clustering held-out training errors and selecting representative examples, TypiClust provides diverse failure patterns that ground LLM reasoning in concrete error magnitudes.

Test-time adaptation provides consistent improvement by detecting distribution shifts and triggering control signal refinement when input statistics diverge. The aggregate effect (+1.03\% MSE) understates its selective impact: on the hardest quartile of ETTh2 samples (Q4 by backbone MSE), the system without adaptation improves by only +1.4\%, while full CTRL reaches +5.7\% (Appendix~\ref{app:case_studies}). Table~\ref{tab:ablation} further confirms each parameter contributes: removing any single dimension degrades performance by 0.48--0.52\%, indicating the decoder utilizes control signal semantics for correction.

LLM choice shows robustness: GPT-4o provides marginal improvement, while smaller models like Qwen 2.5 72B show modest degradation, indicating CTRL is not brittle to LLM selection within capable model families.
\subsection{Computational Efficiency}
\label{sec:efficiency}

CTRL achieves efficiency through lightweight training and minimal LLM inference.
Table~\ref{tab:complexity} compares computational requirements: 
CTRL requires only $\sim$400K trainable parameters (45$\times$ fewer than CALF) 
and trains in under 2 minutes, while the frozen GPT-2 encoder incurs no training cost.

\begin{table}[h!]
\centering
\footnotesize
\setlength{\tabcolsep}{1.5pt}
\begin{tabular}{@{}lccccc@{}}
\toprule
& \textbf{Ours} & \textbf{LLMTim} & \textbf{GPT4TS} & \textbf{CALF} & \textbf{TEMPO} \\
\midrule
Params & 400K & 0 & 4.4M & 18.2M & 12.4M \\
Train & 1.8m & 0 & 2.1m & 62m & 106m \\
LLM & 3--24 & $N$ & 0 & 0 & 0 \\
\bottomrule
\end{tabular}
\caption{Model complexity on ETTh2 (384$\rightarrow$96).}
\label{tab:complexity}
\end{table}

The maximum LLM call count is deterministic given dataset size and calibration interval $K$. CTRL requires 3 initial calls (one per agent) for control signal generation. Test-time adaptation checks every $K{=}50$ batches, adding at most 3 calls per check when distribution shift is detected. Table~\ref{tab:llm_calls} reports the per-dataset breakdown: smaller datasets such as Exchange require only 6 calls, while larger datasets such as ETTm2 require up to 24, yielding over 1,000$\times$ reduction compared to per-sample methods. Sensitivity analysis (Table~\ref{tab:hparam_sensitivity}) confirms that MSE varies less than 0.15\% across calibration intervals from 5 to 100 batches.

\begin{table}[h!]
\centering
\small
\setlength{\tabcolsep}{4pt}
\begin{tabular}{lrrr}
\toprule
Dataset & Test Samples & Checks & Max Calls \\
\midrule
ETTh1/h2 & 2,402 & 2 & 9 \\
ETTm1/m2 & 11,042 & 7 & 24 \\
Weather & 10,060 & 7 & 24 \\
ECL & 4,781 & 3 & 12 \\
Exchange & 1,038 & 1 & 6 \\
\bottomrule
\end{tabular}
\caption{LLM calls per dataset (calibration interval $K{=}50$, $H{=}96$). Total calls $= 3 + 3 \times \text{Checks}$ in the worst case.}
\label{tab:llm_calls}
\end{table}

\section{Conclusion}
\label{sec:conclusion}

We presented CTRL, a framework that repositions LLMs as strategic controllers rather than numerical predictors. Frozen LLMs generate interpretable correction policies via in-context learning, while a lightweight residual decoder translates these into forecast adjustments. Experiments demonstrate consistent gains, with the strongest improvements on non-stationary datasets, using 3--24 LLM calls through policy caching. Structural isolation from pretraining data positions CTRL as a practical plug-in for deployment where distribution shift is unpredictable and labels are unavailable.

\section{Limitations}
CTRL's improvement diminishes at extended horizons, where growing backbone error makes residual correction more uncertain. On highly stationary datasets (e.g., Weather), gains are marginal, which reflects appropriate behavior rather than a failure mode. Our evaluation excludes TimesFM~\cite{das2024timesfm}, a 200M-parameter foundation model representing a different contribution category; applying CTRL as a post-hoc correction layer on foundation model outputs remains future work.

\section*{Acknowledgments}

This work was supported by the Ministry of Education of the Republic of Korea and the National Research Foundation of Korea (NRF-2024S1A5C3A03046579).

\bibliography{ref3}

\begin{thebibliography}{39}
\providecommand{\natexlab}[1]{#1}

\bibitem[{Brown et~al.(2020)Brown, Mann, Ryder, Subbiah, Kaplan, Dhariwal,
  Neelakantan, Shyam, Sastry, Askell et~al.}]{brown2020gpt3}
Tom~B. Brown, Benjamin Mann, Nick Ryder, Melanie Subbiah, Jared Kaplan,
  Prafulla Dhariwal, Arvind Neelakantan, Pranav Shyam, Girish Sastry, Amanda
  Askell, and 1 others. 2020.
\newblock \href {https://doi.org/10.48550/arXiv.2005.14165} {Language models
  are few-shot learners}.
\newblock In \emph{Advances in Neural Information Processing Systems},
  volume~33, pages 1877--1901.

\bibitem[{Cao et~al.(2024)Cao, Jia, Arik, Pfister, Zheng, Ye, and
  Liu}]{Cao2024TEMPO}
Defu Cao, Furong Jia, Sercan~O Arik, Tomas Pfister, Yixiang Zheng, Wen Ye, and
  Yan Liu. 2024.
\newblock \href {https://doi.org/10.48550/arXiv.2310.04948} {{TEMPO}:
  Prompt-based generative pre-trained transformer for time series forecasting}.
\newblock In \emph{International Conference on Learning Representations
  (ICLR)}.

\bibitem[{Chang et~al.(2023)Chang, Peng, and Chen}]{Chang2023LLM4TS}
Ching Chang, Wen-Chih Peng, and Tien-Fu Chen. 2023.
\newblock \href {https://doi.org/10.48550/arXiv.2308.08469} {{LLM4TS}:
  Two-stage fine-tuning for time-series forecasting with pre-trained {LLMs}}.
\newblock \emph{arXiv preprint arXiv:2308.08469}.

\bibitem[{Das et~al.(2024)Das, Kong, Sen, and Zhou}]{das2024timesfm}
Abhimanyu Das, Weihao Kong, Rajat Sen, and Yichen Zhou. 2024.
\newblock A decoder-only foundation model for time-series forecasting.
\newblock In \emph{International Conference on Machine Learning (ICML)}.

\bibitem[{Geng et~al.(2024)Geng, Cai, Wang, Koeppl, Nakov, and
  Gurevych}]{geng2024survey-conf}
Jiahui Geng, Fengyu Cai, Yuxia Wang, Heinz Koeppl, Preslav Nakov, and Iryna
  Gurevych. 2024.
\newblock A survey of confidence estimation and calibration in large language
  models.
\newblock In \emph{Proceedings of the 2024 Conference of the North American
  Chapter of the Association for Computational Linguistics (NAACL)}, pages
  6577--6595.

\bibitem[{Gruver et~al.(2023)Gruver, Finzi, Qiu, and
  Wilson}]{gruver2023llmtime}
Nate Gruver, Marc Finzi, Shikai Qiu, and Andrew~Gordon Wilson. 2023.
\newblock \href {https://doi.org/10.48550/arXiv.2310.07820} {Large language
  models are zero-shot time series forecasters}.
\newblock In \emph{Advances in Neural Information Processing Systems},
  volume~36.

\bibitem[{Hacohen et~al.(2022)Hacohen, Dekel, and
  Weinshall}]{hacohen2022active}
Guy Hacohen, Avihu Dekel, and Daphna Weinshall. 2022.
\newblock \href {https://doi.org/10.48550/arXiv.2202.02794} {Active learning on
  a budget: Opposite strategies suit high and low budgets}.
\newblock In \emph{Proceedings of the 39th International Conference on Machine
  Learning (ICML)}, pages 8175--8195.

\bibitem[{Hochreiter and Schmidhuber(1997)}]{hochreiter1997lstm}
Sepp Hochreiter and J{\"u}rgen Schmidhuber. 1997.
\newblock \href {https://doi.org/10.1162/neco.1997.9.8.1735} {Long short-term
  memory}.
\newblock \emph{Neural Computation}, 9(8):1735--1780.

\bibitem[{Jiang et~al.(2025)Jiang, Yu, Lee, Song, Shin, Cheng, Liu, and
  Chen}]{Jiang2025Explainable}
Yushan Jiang, Wenchao Yu, Geon Lee, Dongjin Song, Kijung Shin, Wei Cheng,
  Yanchi Liu, and Haifeng Chen. 2025.
\newblock \href {https://doi.org/10.48550/arXiv.2503.01013} {{TimeXL}:
  Explainable multi-modal time series prediction with {LLM}-in-the-loop}.
\newblock \emph{arXiv preprint arXiv:2503.01013}.

\bibitem[{Jin et~al.(2024)Jin, Wang, Ma, Chu, Zhang, Shi, Chen, Liang, Li, Pan,
  and Wen}]{Jin2024TimeLLM}
Ming Jin, Shiyu Wang, Lintao Ma, Zhixuan Chu, James~Y Zhang, Xiaoming Shi,
  Pin-Yu Chen, Yuxuan Liang, Yuan-Fang Li, Shirui Pan, and Qingsong Wen. 2024.
\newblock \href {https://doi.org/10.48550/arXiv.2310.01728} {{Time-LLM}: Time
  series forecasting by reprogramming large language models}.
\newblock In \emph{International Conference on Learning Representations
  (ICLR)}.

\bibitem[{Kim et~al.(2022{\natexlab{a}})Kim, Cho, Kim, Park, and
  Choo}]{kim2022rescal}
Daejin Kim, Youngin Cho, Dongmin Kim, Cheonbok Park, and Jaegul Choo.
  2022{\natexlab{a}}.
\newblock \href {https://doi.org/10.1145/3511808.3557432} {Residual correction
  in real-time traffic forecasting}.
\newblock In \emph{Proceedings of the 31st ACM International Conference on
  Information and Knowledge Management (CIKM)}, pages 962--971.

\bibitem[{Kim et~al.(2022{\natexlab{b}})Kim, Kim, Tae, Park, Choi, and
  Choo}]{Kim2022RevIN}
Taesung Kim, Jinhee Kim, Yunwon Tae, Cheonbok Park, Jang-Ho Choi, and Jaegul
  Choo. 2022{\natexlab{b}}.
\newblock \href {https://openreview.net/forum?id=cGDAkQo1C0p} {Reversible
  instance normalization for accurate time-series forecasting against
  distribution shift}.
\newblock In \emph{International Conference on Learning Representations
  (ICLR)}.

\bibitem[{Liang et~al.(2024)Liang, Wen, Nie, Jiang, Jin, Song, Pan, and
  Wen}]{Liang2024Foundation}
Yuxuan Liang, Haomin Wen, Yuqi Nie, Yushan Jiang, Ming Jin, Dongjin Song,
  Shirui Pan, and Qingsong Wen. 2024.
\newblock \href {https://doi.org/10.1145/3637528.3671451} {Foundation models
  for time series analysis: A tutorial and survey}.
\newblock In \emph{Proceedings of the 30th ACM SIGKDD Conference on Knowledge
  Discovery and Data Mining}, pages 6555--6565.

\bibitem[{Liu et~al.(2024{\natexlab{a}})Liu, Xu, Miao, Yang, Zhang, Long, Li,
  and Zhao}]{Liu2025TimeCMA}
Chenxi Liu, Qianxiong Xu, Hao Miao, Sun Yang, Lingzheng Zhang, Cheng Long,
  Ziyue Li, and Rui Zhao. 2024{\natexlab{a}}.
\newblock \href {https://arxiv.org/pdf/2406.01638} {{TimeCMA}: Towards
  {LLM}-empowered multivariate time series forecasting via cross-modality
  alignment}.
\newblock \emph{arXiv preprint arXiv:2406.01638}.

\bibitem[{Liu et~al.(2024{\natexlab{b}})Liu, Yang, Xu, Li, Long, Li, and
  Zhao}]{Liu2024STLLM}
Chenxi Liu, Sun Yang, Qianxiong Xu, Zhishuai Li, Cheng Long, Ziyue Li, and Rui
  Zhao. 2024{\natexlab{b}}.
\newblock \href {https://doi.org/10.48550/arXiv.2401.10134} {Spatial-temporal
  large language model for traffic prediction}.
\newblock \emph{arXiv preprint arXiv:2401.10134}.

\bibitem[{Liu et~al.(2025)Liu, Guo, Dai, Li, Bao, Ren, Jiang, and
  Xia}]{Liu2025CALF}
Peiyuan Liu, Hang Guo, Tao Dai, Naiqi Li, Jigang Bao, Xudong Ren, Yong Jiang,
  and Shu-Tao Xia. 2025.
\newblock \href {https://doi.org/10.1609/aaai.v39i18.34082} {{CALF}: Aligning
  {LLMs} for time series forecasting via cross-modal fine-tuning}.
\newblock In \emph{AAAI Conference on Artificial Intelligence}.

\bibitem[{Liu et~al.(2024{\natexlab{c}})Liu, Hu, Li, Diao, Liang, Hooi, and
  Zimmermann}]{Liu2024UniTime}
Xu~Liu, Junfeng Hu, Yuan Li, Shizhe Diao, Yuxuan Liang, Bryan Hooi, and Roger
  Zimmermann. 2024{\natexlab{c}}.
\newblock \href {https://doi.org/10.1145/3589334.3645434} {{UniTime}: A
  language-empowered unified model for cross-domain time series forecasting}.
\newblock In \emph{Proceedings of the ACM Web Conference (WWW)}, pages
  4095--4106.

\bibitem[{Liu et~al.(2023{\natexlab{a}})Liu, Li, Wang, and Long}]{Liu2023Koopa}
Yong Liu, Chenyu Li, Jianmin Wang, and Mingsheng Long. 2023{\natexlab{a}}.
\newblock \href {https://doi.org/10.48550/arXiv.2305.18803} {Koopa: Learning
  non-stationary time series dynamics with {Koopman} predictors}.
\newblock In \emph{Advances in Neural Information Processing Systems
  (NeurIPS)}, pages 12271--12290.

\bibitem[{Liu et~al.(2024{\natexlab{d}})Liu, Qin, Huang, Wang, and
  Long}]{Liu2024AutoTimes}
Yong Liu, Guo Qin, Xiangdong Huang, Jianmin Wang, and Mingsheng Long.
  2024{\natexlab{d}}.
\newblock \href {https://doi.org/10.52202/079017-3882} {{AutoTimes}:
  Autoregressive time series forecasters via large language models}.
\newblock In \emph{Advances in Neural Information Processing Systems
  (NeurIPS)}.

\bibitem[{Liu et~al.(2023{\natexlab{b}})Liu, Cheng, Li, Huang, Liu, Xie, and
  Chen}]{Liu2023AdaptiveNorm}
Zhiding Liu, Mingyue Cheng, Zhi Li, Zhenya Huang, Qi~Liu, Yanhu Xie, and Enhong
  Chen. 2023{\natexlab{b}}.
\newblock \href {https://doi.org/10.52202/075280-0628} {Adaptive normalization
  for non-stationary time series forecasting: A temporal slice perspective}.
\newblock In \emph{Advances in Neural Information Processing Systems
  (NeurIPS)}, pages 14273--14292.

\bibitem[{Nie et~al.(2023)Nie, Nguyen, Sinthong, and
  Kalagnanam}]{Nie2023PatchTST}
Yuqi Nie, Nam~H. Nguyen, Phanwadee Sinthong, and Jayant Kalagnanam. 2023.
\newblock \href {https://doi.org/10.48550/arXiv.2211.14730} {A time series is
  worth 64 words: Long-term forecasting with transformers}.
\newblock In \emph{International Conference on Learning Representations
  (ICLR)}.

\bibitem[{Niu et~al.(2023)Niu, Wu, Zhang, Wen, Chen, Zhao, and
  Tan}]{niu2023sar}
Shuaicheng Niu, Jiaxiang Wu, Yifan Zhang, Zhiquan Wen, Yaofo Chen, Peilin Zhao,
  and Mingkui Tan. 2023.
\newblock \href {https://doi.org/10.48550/arXiv.2302.12400} {Towards stable
  test-time adaptation in dynamic wild world}.
\newblock In \emph{International Conference on Learning Representations
  (ICLR)}.

\bibitem[{Pan et~al.(2024)Pan, Jiang, Garg, Schneider, Nevmyvaka, and
  Song}]{Pan2024S2IP}
Zijie Pan, Yushan Jiang, Sahil Garg, Anderson Schneider, Yuriy Nevmyvaka, and
  Dongjin Song. 2024.
\newblock \href {https://doi.org/10.48550/arXiv.2403.05798} {{S$^{2}$IP-LLM}:
  Semantic space informed prompt learning with {LLM} for time series
  forecasting}.
\newblock In \emph{International Conference on Machine Learning (ICML)}.

\bibitem[{Park et~al.(2025)Park, Lee, Lee, Gwak, and Choo}]{park2025revisiting}
Junwoo Park, Hyuck Lee, Dohyun Lee, Daehoon Gwak, and Jaegul Choo. 2025.
\newblock \href {https://doi.org/10.18653/v1/2025.acl-short.71} {Revisiting
  {LLM}s as zero-shot time series forecasters: Small noise can break large
  models}.
\newblock In \emph{Proceedings of the 63rd Annual Meeting of the Association
  for Computational Linguistics (Volume 2: Short Papers)}, pages 906--922,
  Vienna, Austria. Association for Computational Linguistics.

\bibitem[{Perez et~al.(2018)Perez, Strub, de~Vries, Dumoulin, and
  Courville}]{perez2018film}
Ethan Perez, Florian Strub, Harm de~Vries, Vincent Dumoulin, and Aaron
  Courville. 2018.
\newblock \href {https://doi.org/10.1609/aaai.v32i1.11671} {{FiLM}: Visual
  reasoning with a general conditioning layer}.
\newblock In \emph{Proceedings of the AAAI Conference on Artificial
  Intelligence}.

\bibitem[{Schick et~al.(2023)Schick, Dwivedi-Yu, Dess{\`i}, Raileanu, Lomeli,
  Zettlemoyer, Cancedda, and Scialom}]{schick2023toolformer}
Timo Schick, Jane Dwivedi-Yu, Roberto Dess{\`i}, Roberta Raileanu, Maria
  Lomeli, Luke Zettlemoyer, Nicola Cancedda, and Thomas Scialom. 2023.
\newblock \href {https://doi.org/10.48550/arXiv.2302.04761} {Toolformer:
  Language models can teach themselves to use tools}.
\newblock In \emph{Advances in Neural Information Processing Systems},
  volume~36.

\bibitem[{Singh and Strouse(2024)}]{Singh2024TokenizationCounts}
Aaditya~K. Singh and DJ~Strouse. 2024.
\newblock \href {https://doi.org/10.48550/arXiv.2402.14903} {Tokenization
  counts: The impact of tokenization on arithmetic in frontier {LLMs}}.
\newblock \emph{arXiv preprint arXiv:2402.14903}.

\bibitem[{Sun et~al.(2024)Sun, Li, Li, and Hong}]{Sun2023TEST}
Chenxi Sun, Hongyan Li, Yaliang Li, and Shenda Hong. 2024.
\newblock \href {https://doi.org/10.48550/arXiv.2308.08241} {{TEST}: Text
  prototype aligned embedding to activate {LLM's} ability for time series}.
\newblock In \emph{International Conference on Learning Representations
  (ICLR)}.

\bibitem[{Tan et~al.(2024)Tan, Merrill, Gupta, Althoff, and
  Hartvigsen}]{Tan2024LLMUseful}
Mingtian Tan, Mike~A. Merrill, Vinayak Gupta, Tim Althoff, and Thomas
  Hartvigsen. 2024.
\newblock \href {https://doi.org/10.52202/079017-1922} {Are language models
  actually useful for time series forecasting?}
\newblock In \emph{Advances in Neural Information Processing Systems
  (NeurIPS)}.

\bibitem[{Wang et~al.(2021)Wang, Shelhamer, Liu, Olshausen, and
  Darrell}]{wang2021tent}
Dequan Wang, Evan Shelhamer, Shaoteng Liu, Bruno Olshausen, and Trevor Darrell.
  2021.
\newblock \href {https://doi.org/10.48550/arXiv.2006.10726} {Tent: Fully
  test-time adaptation by entropy minimization}.
\newblock In \emph{International Conference on Learning Representations
  (ICLR)}.

\bibitem[{Wei et~al.(2022)Wei, Wang, Schuurmans, Bosma, Ichter, Xia, Chi, Le,
  and Zhou}]{wei2022cot}
Jason Wei, Xuezhi Wang, Dale Schuurmans, Maarten Bosma, Brian Ichter, Fei Xia,
  Ed~Chi, Quoc~V Le, and Denny Zhou. 2022.
\newblock \href {https://doi.org/10.48550/arXiv.2201.11903} {Chain-of-thought
  prompting elicits reasoning in large language models}.
\newblock In \emph{Advances in Neural Information Processing Systems},
  volume~35, pages 24824--24837.

\bibitem[{Xiong et~al.(2024)Xiong, Hu, Lu, Li, Fu, He, and
  Hooi}]{xiong2024llms-conf}
Miao Xiong, Zhiyuan Hu, Xinyang Lu, Yifei Li, Jie Fu, Junxian He, and Bryan
  Hooi. 2024.
\newblock \href {https://doi.org/10.48550/arXiv.2306.13063} {Can {LLM}s express
  their uncertainty? {An} empirical evaluation of confidence elicitation in
  {LLMs}}.
\newblock In \emph{International Conference on Learning Representations
  (ICLR)}.

\bibitem[{Xue and Salim(2023)}]{Xue2023PromptCast}
Hao Xue and Flora~D. Salim. 2023.
\newblock \href {https://doi.org/10.1109/tkde.2023.3342137} {{PromptCast}: A
  new prompt-based learning paradigm for time series forecasting}.
\newblock \emph{IEEE Transactions on Knowledge and Data Engineering}, pages
  6851--6864.

\bibitem[{Yao et~al.(2023)Yao, Zhao, Yu, Du, Shafran, Narasimhan, and
  Cao}]{yao2023react}
Shunyu Yao, Jeffrey Zhao, Dian Yu, Nan Du, Izhak Shafran, Karthik Narasimhan,
  and Yuan Cao. 2023.
\newblock \href {https://doi.org/10.48550/arXiv.2210.03629} {{ReAct}:
  Synergizing reasoning and acting in language models}.
\newblock In \emph{International Conference on Learning Representations
  (ICLR)}.

\bibitem[{Zeng et~al.(2023)Zeng, Chen, Zhang, and Xu}]{Zeng2023DLinear}
Ailing Zeng, Muxi Chen, Lei Zhang, and Qiang Xu. 2023.
\newblock \href {https://doi.org/10.1609/aaai.v37i9.26317} {Are transformers
  effective for time series forecasting?}
\newblock In \emph{Proceedings of the AAAI Conference on Artificial
  Intelligence}.

\bibitem[{Zhang et~al.(2025)Zhang, Han, Fang, Ansari, Zhang, Maddix, Hu,
  Wilson, Mahoney, Wang et~al.}]{Zhang2025DoesMultimodality}
Xiyuan Zhang, Boran Han, Haoyang Fang, Abdul~Fatir Ansari, Shuai Zhang,
  Danielle~C. Maddix, Cuixiong Hu, Andrew~Gordon Wilson, Michael~W. Mahoney,
  Hao Wang, and 1 others. 2025.
\newblock \href {https://doi.org/10.48550/arXiv.2506.21611} {Does multimodality
  lead to better time series forecasting?}
\newblock arXiv preprint arXiv:2506.21611.

\bibitem[{Zhang et~al.(2023)Zhang, Wen, Wang, Chen, Sun, Zhang, Wang, Jin, and
  Tan}]{Wen2023OneNet}
Yi-Fan Zhang, Qingsong Wen, Xue Wang, Weiqi Chen, Liang Sun, Zhang Zhang, Liang
  Wang, Rong Jin, and Tieniu Tan. 2023.
\newblock {OneNet}: Enhancing time series forecasting models under concept
  drift by online ensembling.
\newblock In \emph{Advances in Neural Information Processing Systems
  (NeurIPS)}, pages 69949--69980.

\bibitem[{Zhou et~al.(2023)Zhou, Niu, Wang, Sun, and Jin}]{zhou2023onefitsall}
Tian Zhou, Peisong Niu, Xue Wang, Liang Sun, and Rong Jin. 2023.
\newblock \href {https://doi.org/10.48550/arXiv.2302.11939} {{One Fits All}:
  Power general time series analysis by pretrained {LM}}.
\newblock In \emph{Advances in Neural Information Processing Systems},
  volume~36.

\bibitem[{Zhou et~al.(2025)Zhou, Wang, Qu, Zhang, and
  Bergmeir}]{Zhou2025Unveiling}
Xin Zhou, Weiqing Wang, Shilin Qu, Zhiqiang Zhang, and Christoph Bergmeir.
  2025.
\newblock \href {https://doi.org/10.48550/arXiv.2501.07048} {Unveiling the
  potential of text in high-dimensional time series forecasting}.
\newblock \emph{arXiv preprint arXiv:2501.07048}.

\end{thebibliography}

\newpage

\appendix
\raggedbottom
\section*{Appendix}

\section{Implementation Details}
\label{sec:imple}

\subsection{Dataset Statistics}

\begin{table}[H]
\centering
\footnotesize

\resizebox{\columnwidth}{!}{%
\setlength{\tabcolsep}{3pt}
\begin{tabular}{lrrrrl}
\toprule
Dataset & Total & Train & Val & Test & Freq. \\
\midrule
ETTh1/h2 & 17,420 & 12,194 & 1,742 & 3,484 & Hourly \\
ETTm1/m2  & 69,680 & 48,776 & 6,968 & 11,042 & 15-min \\ 
Weather & 52,696 & 36,887 & 5,269 & 10,540 & 10-min \\
ECL & 26,304 & 18,412 & 2,631 & 5,261 & Hourly \\
Exchange & 7,588 & 5,311 & 759 & 1,518 & Daily \\
\bottomrule
\end{tabular}
}
\caption{Dataset statistics and split configuration.}
\label{tab:datasets}
\end{table}

\subsection{Hyperparameters}
\begin{table}[H]
\centering
\resizebox{\columnwidth}{!}{%
\begin{tabular}{l|ccc}
\toprule
Dataset & LR & Epochs  & STL (Seas., Trend) \\
\midrule
ETTh1 & 1e-4 & 40 & 24, 48 \\
ETTh2 & 1e-4 & 40 & 24, 48 \\
ETTm1 & 1e-4 & 40 & 24, 48 \\
ETTm2 & 1e-4 & 40 & 24, 48 \\
ECL & 5e-6 & 40 & 12, 24 \\
Exchange & 1e-4 & 40 & 7, 24 \\
Weather & 3e-5 & 40 & 7, 48 \\
\bottomrule
\end{tabular}%
}
\vspace{-2mm}
{\footnotesize $^\dagger$Weight decay=1e-7, LLM temp=0.3, top-$p$=0.8 for all.}
\caption{Dataset-specific hyperparameters.$^\dagger$}
\label{tab:hyperparams}
\end{table}

Table~\ref{tab:hyperparams} summarizes dataset-specific hyperparameters for decoder training, STL decomposition, and LLM sampling. The residual decoder is a 2-layer MLP with hidden dimension $L \times F$, scaling capacity with input complexity.
The text projection dimension for the Irregular Agent embedding is set to 32.
All experiments were executed on a system with an NVIDIA RTX 3080 GPU, Intel Xeon E5-2686 v4 CPU (2.30 GHz), and 128GB RAM. Code is available at \url{https://github.com/mfriendly/CTRL}.

\subsection{STL Context for Control Signal Generation}
\label{app:stl_context}
For each sample, agents receive historical (input sequence) and prediction (backbone output) statistics:
\begin{equation}
    \mathcal{C} = \{(\mathcal{C}_\text{hist}, \mathcal{C}_\text{pred}) \mid \mu, \sigma, \text{slope}_T, \text{amp}_S, d_S\}
\end{equation}
where $\mu$ and $\sigma$ denote mean and standard deviation, $\text{slope}_T$ is the trend slope via linear regression, $\text{amp}_S$ is the seasonal amplitude (max $-$ min), and $d_S$ is seasonal dominance (ratio of seasonal variance to total variance). This dual-view enables diagnosis by comparing expected versus predicted temporal characteristics.

\subsection{Shift Detection}
\label{app:shift_detection}

We quantify distribution shift separately for each STL component using z-score normalization against held-out training statistics.

Trend shift captures slope changes:
\begin{equation}
s_\text{trend} = \frac{|\text{slope}_\text{test} - \bar{\text{slope}}_\text{val}|}{\sigma_{\text{slope,val}} + \epsilon}
\end{equation}

Seasonal shift measures dominance changes:
\begin{equation}
s_\text{seasonal} = \frac{|d_{S,\text{test}} - \bar{d}_{S,\text{val}}|}{\sigma_{d_S,\text{val}} + \epsilon}
\end{equation}
where seasonal dominance \textit{$d_S = \text{Var}(\text{seasonal}) / \text{Var}(\text{total})$}.

Irregular shift tracks noise level:
\begin{equation}
s_\text{irregular} = \frac{|\sigma_\text{test} - \bar{\sigma}_\text{val}|}{\sigma_{\sigma,\text{val}} + \epsilon}
\end{equation}

Adaptation activates when $s > \tau$ (default $\tau=2.0$), corresponding to test statistics exceeding two standard deviations from the held-out training distribution. The component-wise formulation allows targeted adjustment: a trend reversal need not affect seasonal correction.

\subsection{Hyperparameter Sensitivity}

\begin{table}[H]

\centering
\small
\begin{tabular}{lc}
\toprule
\textbf{Configuration} & $\Delta$\textbf{MSE (\%)} \\
\midrule
\multicolumn{2}{l}{\textit{LLM Sampling (Llama 3.3 70B)}} \\
\quad temp=0.2, top-p=0.8 & +0.22 \\
\quad \textbf{temp=0.3, top-p=0.8} & \textbf{--} \\
\quad temp=0.5, top-p=0.8 & +0.26 \\
\midrule
\multicolumn{2}{l}{\textit{Adaptation Threshold ($\tau$)}} \\
\quad $\tau$=OFF (no adaptation) & +1.03 \\
\quad $\tau$=1.5 & +0.05 \\
\quad $\boldsymbol{\tau}$\textbf{=2.0} & \textbf{--} \\
\quad $\tau$=2.5 & +0.01 \\
\bottomrule
\end{tabular}

\caption{Hyperparameter sensitivity on ETTh2 (384$\to$96). Bold indicates selected configuration. $\Delta$ shows degradation relative to selected setting.}
\label{tab:hparam_sensitivity}
\end{table}

Table~\ref{tab:hparam_sensitivity} evaluates CTRL's sensitivity to key hyperparameters on ETTh2. LLM sampling parameters (temperature, top-p) have minimal impact across the tested range, indicating that control signal generation is stable across reasonable sampling configurations. The adaptation threshold $\tau$ shows an optimum at 1.5--2.5, corresponding to the z-score formulation where adaptation triggers for test statistics exceeding two standard deviations from held-out training. Disabling adaptation entirely degrades performance by 1.03\%.

\section{Efficiency Analysis}
\label{app:efficiency}

\subsection{LLM Call Analysis.}
Test-time adaptation monitors distribution shifts at fixed intervals rather than per-sample, with total calls equal to $3 + 3K$ in the worst case where all $K$ checks trigger adaptation (Table~\ref{tab:llm_calls}). Per-sample policies yield only $<$1\% MSE improvement at 200$\times$ cost compared to shared policies, and varying the calibration interval from 5 to 100 batches changes MSE by under 0.15\% across all six datasets (identical on ETTh1). This confirms that Phase 1's initial control signal generation captures the primary distribution gap, while Phase 3 provides marginal refinement, allowing coarse-grained monitoring (default: 50 batches) to reduce LLM calls by 10--50$\times$ without accuracy loss.

\subsection{Model Complexity Breakdown.}
Table~\ref{tab:param_breakdown} provides detailed parameter counts.
The frozen GPT-2 encoder (124M parameters) adds no training cost; 
only the lightweight MLP projector (100K) and residual decoder (215K) are trained.

\begin{table}[H]
\centering

\small
\begin{tabular}{p{4.5cm}p{1.2cm}}
\toprule
\textbf{Component} & \textbf{Params} \\
\midrule
DLinear backbone (frozen) & 74K \\
GPT-2 encoder (frozen) & 124M \\
MLP projector (trained) & 100K \\
Residual decoder wo proj. (trained) & 215K \\
\midrule
\textbf{Total trainable} & \textbf{$\sim$400K} \\
\bottomrule
\end{tabular}
\caption{CTRL parameter breakdown (ETTh2, DLinear backbone).}
\label{tab:param_breakdown}
\end{table}

\section{Case Studies}
\label{app:case_studies}

\subsection{Quantitative: Ablation Analysis}
\label{app:ablation_analysis}

\begin{figure*}[h!]
    \centering
    \includegraphics[width=\linewidth]{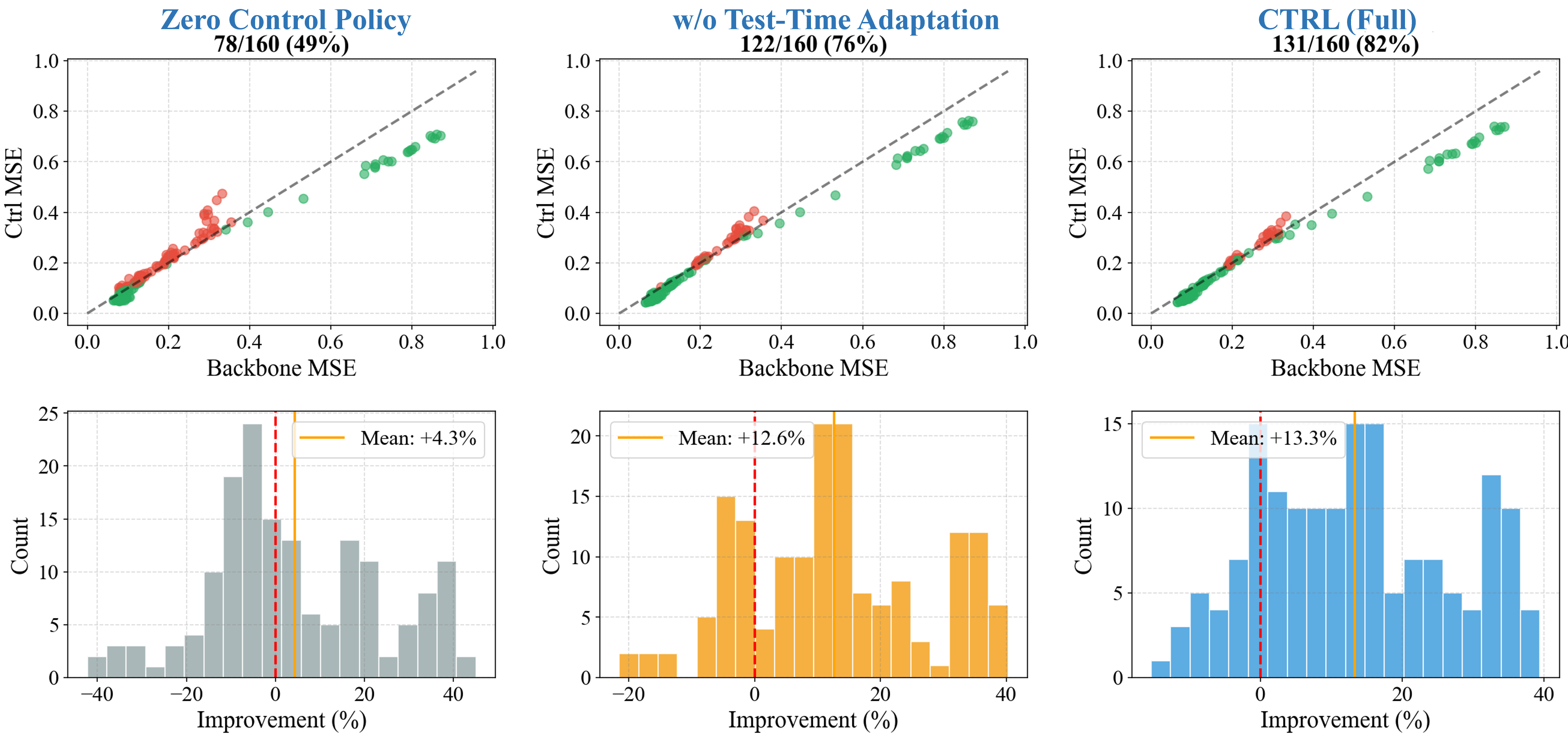}
\caption{Three-way ablation on ETTh2. Scatter plots (top): zero control signal clusters around the diagonal with near-random win rate (49\%), indicating the decoder alone learns minimal correction. Adding LLM control signals shifts points below the diagonal, but gains concentrate in low-to-mid MSE regions. The full system extends improvements to harder samples where distribution shift invalidates cached control signals. Histograms (bottom): test-time adaptation compresses negative-skewed tails by detecting when cached control signals fail and triggering re-reasoning.}
    \label{fig:3way_ablation}
\end{figure*}

We conduct a three-way ablation to isolate contributions of each CTRL component: (1) Zero Control Signal (decoder only, null control signal), (2) w/o Test-Time Adaptation (LLM-generated control signal only), and (3) CTRL Full (control signal generation + test-time adaptation). Results are evaluated on ETTh2 (384$\to$96) with DLinear backbone.

\begin{table}[h!]
\centering
\footnotesize
\begin{tabular}{lrrr}
\toprule
& \textbf{Zero} & \textbf{w/o TTA} & \textbf{Full} \\
\midrule
Mean Imp. & +4.3\% & +12.6\% & +13.3\% \\
Win Rate & 49\% & 76\% & 82\% \\
\bottomrule
\end{tabular}
\caption{Three-way ablation on ETTh2.}
\label{tab:3way_ablation}
\end{table}

\begin{table}[h!]
\centering
\small
\footnotesize
\begin{tabular}{lc}
\toprule
\textbf{Comparison} & \textbf{$\Delta$ Improv.} \\
\midrule
w/o Adaptation vs Zero & +8.3\% \\
CTRL (Full) vs w/o Adaptation & +0.7\% \\
CTRL (Full) vs Zero & +9.0\% \\
\bottomrule
\end{tabular}
\caption{Relative contribution of each component.}
\label{tab:3way_contributions}
\end{table}

Table~\ref{tab:3way_ablation} quantifies overall performance, but the aggregate metrics obscure where each component provides value. The near-random win rate (49\%) under zero control signal confirms that without LLM guidance, the residual decoder lacks sufficient inductive bias to systematically correct backbone errors; the decoder architecture alone cannot discover correction strategies from data.

The 27\% win rate jump when adding LLM control signals (49\%$\to$76\%) reflects LLM reasoning's capacity to diagnose backbone failure modes from held-out training statistics and encode correction strategies into compact control signals. However, Figure~\ref{fig:3way_ablation} reveals these gains concentrate disproportionately on easier samples (Q1-Q2), where backbone errors follow patterns visible in held-out training data. On harder samples (Q3-Q4), improvements without adaptation diminish because the control signal assumes distributional stationarity that does not hold.

Test-time adaptation provides the critical bridge to hard samples. While its aggregate contribution appears modest (+0.7\% mean improvement, +6\% win rate), this masks its selective impact: on Q4 samples where the system without adaptation shows only +1.4\% improvement, CTRL (Full) achieves +5.7\%. The mechanism is straightforward: by monitoring STL statistics at test time, CTRL detects when input distributions diverge from held-out training and triggers re-reasoning rather than applying stale control signals. This explains the histogram compression in Figure~\ref{fig:3way_ablation} (bottom right), where the negative tail from hard-sample failures shrinks under full adaptation.

The decomposition in Table~\ref{tab:3way_contributions} thus understates test-time adaptation's role. LLM control signal generation accounts for the majority of aggregate improvement (+8.3\% of +9.0\%), but test-time adaptation determines whether CTRL degrades gracefully or catastrophically under distribution shift. For deployment scenarios where test distributions are unpredictable, this robustness property may matter more than average-case gains.

\subsection{Qualitative: Agent Reasoning Traces}

We examine two contrasting cases from ETTh2 (384$\rightarrow$96) to illustrate how test-time adaptation responds to different shift patterns. We select samples exhibiting distinct shift patterns: Sample 0 (highest trend shift score in batch 0) and Sample 640 (highest seasonal shift score in batch 1).

\subsubsection{Case 1: Trend Reversal}

The trend slope reverses from $+3.05 \times 10^{-5}$ (held-out training) to $-1.57 \times 10^{-5}$ (test), yielding a shift score of 1.51 (151\%). Seasonal dominance increases moderately from 0.395 to 0.486 (23\% shift), while noise level rises slightly from $\sigma=1.07$ to $\sigma=1.19$ (11\% shift). The held-out training control signal assumed a rising trend; applying it to declining test data would systematically overpredict.

\begin{figure}[h!]
\centering
\fbox{\parbox{0.95\columnwidth}{
\textbf{Trend Agent} \\
\textit{Diagnosis:} The validation and test trends have a significant shift in direction with a ratio of 151.33\%, indicating a reversal in trend. \\
\textit{Reasoning:} Reduced scale multiplier to dampen the impact of extreme predictions, negative bias shift to adjust for the reversal in trend direction, increased gate multiplier to allow more flexibility in response to new data, and a confidence penalty to reflect reduced certainty due to the significant trend change.
}}
\caption{Trend agent reasoning for Case 1 (trend reversal).}
\label{fig:case1_trend}
\end{figure}

\begin{figure}[h!]
\centering
\fbox{\parbox{0.95\columnwidth}{
\textbf{Seasonal Agent} \\
\textit{Diagnosis:} The test data shows a stronger seasonal pattern compared to the validation data with a significant shift ratio. \\
\textit{Reasoning:} Adjustments are made to amplify the seasonal scale and gate multipliers to better capture the increased strength of the seasonal pattern in test data, while a slight bias shift is added to align predictions with observed trends. Confidence penalty reduces overconfidence due to model adjustments.
}}
\caption{Seasonal agent reasoning for Case 1.}
\label{fig:case1_seasonal}
\end{figure}

Table~\ref{tab:case1_policy} shows the resulting numerical control signal changes. Trend parameters shift dramatically (scale halved from 0.90 to 0.50, confidence reduced from 0.60 to 0.21), while seasonal scale increases from 0.95 to 1.50 to amplify the strengthening pattern. Gates increase across all components (0.40$\rightarrow$0.69 for trend), reflecting higher uncertainty where the LLM delegates more correction authority to the learned decoder.

\begin{table}[h!]
\centering
\small

\begin{tabular}{lcccc}
\toprule
& Scale & Bias & Gate & Conf. \\
\midrule
\multicolumn{5}{l}{\textit{Before Adaptation (Stage 1)}} \\
Trend & 0.90 & $+$0.10 & 0.40 & 0.60 \\
Seasonal & 0.95 & $+$0.10 & 0.40 & 0.60 \\
Irregular & 0.95 & $+$0.10 & 0.50 & 0.60 \\
\midrule
\multicolumn{5}{l}{\textit{After Adaptation (Stage 3)}} \\
Trend & 0.50 & $-$0.50 & 0.69 & 0.21 \\
Seasonal & 1.50 & $+$0.55 & 0.66 & 0.41 \\
Irregular & 1.26 & $-$0.20 & 0.67 & 0.31 \\
\bottomrule
\end{tabular}
\caption{Control signal values before and after adaptation (Case 1: Trend Reversal).}
\label{tab:case1_policy}
\end{table}

\subsubsection{Case 2: Seasonal Amplification (Sample 640)}

Here the pattern inverts: trend shift is minimal (score 0.02, slope unchanged at $\approx 3.0 \times 10^{-5}$), but seasonal dominance more than doubles from 0.395 to 0.849 (115\% shift). The test window falls in a period where periodic fluctuations overpower the stable trend.

\begin{figure}[h!]
\centering
\fbox{\parbox{0.95\columnwidth}{
\textbf{Trend Agent} \\
\textit{Diagnosis:} A small positive shift in the trend slope has been observed with a minimal change ratio of 2.24\%, indicating a slight increase in the rate of change without any direction reversal. \\
\textit{Reasoning:} Adjustments are made to slightly increase the scale and gate multipliers to account for the positive shift in trend slope, while a small bias shift is applied to align predictions more closely with observed patterns.
}}
\caption{Trend agent reasoning for Case 2 (seasonal amplification).}
\label{fig:case2_trend}
\end{figure}

\begin{figure}[h!]
\centering
\fbox{\parbox{0.95\columnwidth}{
\textbf{Seasonal Agent} \\
\textit{Diagnosis:} The test data shows a significantly stronger seasonal pattern compared to the validation data. \\
\textit{Reasoning:} Increasing the scale multiplier and gate multiplier to account for the stronger seasonal pattern in test data, adding a bias shift to further adjust predictions upwards, and applying a confidence penalty to reflect increased uncertainty in seasonal predictions due to the significant pattern change.
}}
\caption{Seasonal agent reasoning for Case 2.}
\label{fig:case2_seasonal}
\end{figure}

Table~\ref{tab:case2_policy} shows the contrasting adaptation. Unlike Case 1, trend parameters change minimally (scale 0.90$\rightarrow$0.58, moderate adjustment). The seasonal agent, detecting strong amplification, maximizes scale to 1.50 and increases bias to $+$0.60 to match the new dominance level. This demonstrates context-sensitivity: the same adaptation mechanism produces different adjustments depending on which component exhibits shift.

\begin{table}[h!]
\centering
\small

\begin{tabular}{lcccc}
\toprule
& Scale & Bias & Gate & Conf. \\
\midrule
\multicolumn{5}{l}{\textit{Before Adaptation (Stage 1)}} \\
Trend & 0.90 & $+$0.10 & 0.40 & 0.60 \\
Seasonal & 0.95 & $+$0.10 & 0.40 & 0.60 \\
Irregular & 0.95 & $+$0.10 & 0.50 & 0.60 \\
\midrule
\multicolumn{5}{l}{\textit{After Adaptation (Stage 3)}} \\
Trend & 0.58 & $-$0.29 & 0.59 & 0.28 \\
Seasonal & 1.50 & $+$0.60 & 0.61 & 0.42 \\
Irregular & 1.25 & $-$0.14 & 0.63 & 0.37 \\
\bottomrule
\end{tabular}
\caption{Control signal values before and after adaptation (Case 2: Seasonal Amplification).}
\label{tab:case2_policy}
\end{table}

\subsubsection{Design Rationale}

These cases illustrate several key properties of CTRL's test-time adaptation:

\textbf{Component-wise independence.} Each STL component adapts independently, preventing a large trend shift from disrupting well-calibrated seasonal correction. In Case 1, the extreme trend reversal triggers aggressive trend dampening while seasonal adjustment remains moderate.

\textbf{Context-sensitive responses.} The same adaptation mechanism produces opposite adjustments depending on shift direction. Case 1 triggers conservative trend scaling (0.50) with high gate delegation (0.69), while Case 2 triggers aggressive seasonal amplification (1.50) with moderate gate (0.61).

\textbf{Uncertainty-aware delegation.} Confidence parameters absorb uncertainty: high shift scores drop confidence substantially (0.60$\rightarrow$0.21 in Case 1 trend), signaling to the decoder that the LLM is less certain. Simultaneously, gates increase to delegate more correction authority to the learned decoder, implementing a soft handoff under distribution shift.

\textbf{Label-free operation.} Adaptation uses only input statistics from STL decomposition of the lookback window, requiring no ground truth labels. This enables deployment in streaming settings where labels arrive delayed or never.

\subsubsection{Irregular Agent Output}

Unlike trend and seasonal agents that output numerical control signals, the irregular agent generates natural language encoded via frozen GPT-2 and projected to a 32D embedding. Figure~\ref{fig:irregular_output} shows the output format from ETTh2.

\begin{figure}[h!]
\centering
\fbox{\parbox{0.95\columnwidth}{
\textbf{Irregular Agent Output} \\[0.5em]
\textit{[CoT Reasoning]} \textbf{Pattern Recognition:} Historical data shows a slight downward trend with high seasonality. The values fluctuate but generally decrease over time, indicating cyclical behavior. \textbf{Backbone Assessment:} Backbone predictions are slightly off as they predict a stronger downward trend than historical data suggests. They under-predict the mean and volatility compared to historical patterns. \textbf{Uncertainty Consideration:} Confidence should be moderate due to high residual noise (0.2549) indicating significant unexplained variation. The heavy tail difference in kurtosis suggests occasional extreme values. \textbf{Held-out Training Context:} Mean error of 0.0510 and standard deviation of 0.4510, suggesting the backbone under-predicts slightly. \\[0.5em]
\textit{[Reflection]} Given the historical data's slight downward trend and high seasonality, we should DAMPEN backbone predictions (scale $<$ 1.0). A POSITIVE bias is needed since the model under-predicts slightly. A MODERATE gate ($\approx$ 0.5) balances adjustment and uncertainty. \\[0.5em]
\textit{[Control Signal]} Dampen slightly, add positive bias, apply moderate correction strength.
}}
\caption{Chain-of-thought output from irregular agent on ETTh2.}
\label{fig:irregular_output}
\end{figure}

We use text for the irregular component because, unlike trend (characterized by slope) or seasonality (period and amplitude), irregularity lacks a compact parametric description. Phrases like ``high residual noise (0.2549)'' and ``heavy tail difference in kurtosis'' convey nuance that a 4D vector cannot capture. The embedding allows the decoder to learn associations between linguistic descriptions and appropriate corrections.

\section{Control Signal Sensitivity Analysis}
\label{app:sensitivity}

To quantify the decoder's dependence on each control signal component, we perform a parameter sweep analysis on ETTh2 (DLinear, 384$\to$96). For each parameter, we vary its value across the full operational range while holding all other parameters fixed at their LLM-generated values, and measure the mean absolute change in decoder correction output:
\begin{equation}
    \Delta_c = \max_{v \in \mathcal{V}_c} \frac{1}{N} \sum_{i=1}^{N} \left| g_\phi(\hat{y}_i, z) - g_\phi(\hat{y}_i, z_{c \leftarrow v}) \right|
\end{equation}
where $z_{c \leftarrow v}$ denotes the control signal with component $c$ set to sweep value $v$, and $\mathcal{V}_c$ is the operational range for that component. For numeric parameters, we sweep: scale $\in [0.5, 1.5]$, bias $\in [-1.0, 1.0]$, gate $\in [0.0, 1.0]$, confidence $\in [0.0, 1.0]$. For the irregular text embedding, we measure the output change when replacing the LLM-generated text with an empty string, removing the 32-dimensional projected embedding entirely. We use $N = 100$ uniformly sampled test windows.

\begin{table}[H]
\centering
\small
\begin{tabular}{lc}
\toprule
\textbf{Component} & $\boldsymbol{\Delta}$ \textbf{(ETTh2)} \\
\midrule
\textbf{Irregular text emb.} & \textbf{0.032} \\
Trend bias & 0.009 \\
Trend confidence & 0.006 \\
Trend scale & 0.005 \\
Trend gate & 0.005 \\
Seasonal confidence & 0.001 \\
Seasonal scale & 0.001 \\
Seasonal gate & 0.001 \\
Seasonal bias & 0.000 \\
\bottomrule
\end{tabular}
\caption{Control signal sensitivity on ETTh2 (384$\to$96, DLinear). The irregular text embedding is 3.4$\times$ more influential than the most responsive numeric parameter (trend bias).}
\label{tab:sensitivity}
\end{table}

Table~\ref{tab:sensitivity} confirms that the decoder has learned to leverage the 32-dimensional GPT-2 projected text representation as its primary source of semantic context. We additionally test how different text inputs affect the decoder output (Table~\ref{tab:text_variants}), replacing the LLM-generated text with various alternatives.

\begin{table}[H]
\centering
\small
\begin{tabular}{lc}
\toprule
\textbf{Text Variant} & $\boldsymbol{\Delta}$ \\
\midrule
No text (zeroed embedding) & 0.0323 \\
Random unrelated words & 0.0074 \\
Technical jargon & 0.0045 \\
Crisis/volatility description & 0.0010 \\
Random characters & 0.0005 \\
Calm/stable description & 0.0004 \\
\bottomrule
\end{tabular}
\caption{Decoder output change when replacing LLM-generated irregular text with alternative inputs on ETTh2.}
\label{tab:text_variants}
\end{table}

The large gap between removing the text entirely ($\Delta = 0.032$) and substituting alternative texts ($\Delta \leq 0.007$) indicates that the decoder distinguishes the \emph{presence} of the LLM text embedding from its \emph{absence}, while the variation among different text inputs confirms sensitivity to \emph{content}. Combined with the random control signal ablation (Table~\ref{tab:ablation}), which shows random text through the same GPT-2 pipeline performs at the zero-signal level, these results confirm that the decoder utilizes semantic content from the LLM rather than treating the embedding as a generic bias term. This text variant analysis serves as a functional probe of the irregular embedding space: 
by systematically substituting inputs and measuring output divergence, we directly test 
whether the decoder has learned to extract semantic content from the embedding rather 
than relying on its statistical properties alone.

\section{Prompts}
\label{sec:appendix_prompts}

\subsection{Multi-Agent Control Signal Generation}

CTRL employs three specialized agents for STL component-aware control signal generation. Each agent receives multi-scale context (global statistics, few-shot examples, current STL comparison) and outputs structured control signals.

\begin{table}[h!]
\centering
\scriptsize
\setlength{\tabcolsep}{3pt}
\begin{tabular}{p{1.8cm}p{5.2cm}}
\toprule
\multicolumn{2}{l}{\textbf{Trend Specialist Agent}} \\
\midrule
Role & First specialist handling TREND correction \\
Focus & Low-frequency patterns: direction, level shifts, drift, slope accuracy \\
Key Questions & (1) Slope sign match? (2) Slope value match? (3) Systematic over/under-prediction? \\
\midrule
\multicolumn{2}{l}{\textit{Output: 4D control signal}} \\
\texttt{scale} & Trend amplitude adjustment ($>$1 if underestimate, $<$1 if overestimate) \\
\texttt{bias} & Level correction ($+$ if too low, $-$ if too high) \\
\texttt{gate} & Correction aggressiveness \\
\texttt{confidence} & Agent certainty \\
\midrule
\multicolumn{2}{l}{\textit{Dynamic context provided}} \\
Comparison & Historical vs backbone: direction, slope, strength \\
\bottomrule
\end{tabular}
\caption{Trend Agent prompt structure.}
\label{tab:prompt_trend}
\end{table}

\begin{table}[h!]
\centering
\scriptsize
\setlength{\tabcolsep}{3pt}
\begin{tabular}{p{1.8cm}p{5.2cm}}
\toprule
\multicolumn{2}{l}{\textbf{Seasonal Specialist Agent}} \\
\midrule
Role & Second specialist handling SEASONAL correction \\
Focus & High-frequency periodic patterns: daily/weekly cycles, amplitude, phase \\
Key Questions & (1) Correct periodicity? (2) Amplitude correct? (3) Phase aligned? \\
\midrule
\multicolumn{2}{l}{\textit{Output: 4D control signal}} \\
\texttt{scale} & Amplitude adjustment ($>$1 if dampened, $<$1 if exaggerated) \\
\texttt{bias} & Baseline shift (usually $\approx 0$) \\
\texttt{gate} & Correction aggressiveness \\
\texttt{confidence} & Agent certainty \\
\midrule
\multicolumn{2}{l}{\textit{Dynamic context provided}} \\
Comparison & Historical vs backbone: amplitude, dominance, period \\
\bottomrule
\end{tabular}
\caption{Seasonal Agent prompt structure.}
\label{tab:prompt_seasonal}
\end{table}

\begin{table}[h!]
\centering
\scriptsize
\setlength{\tabcolsep}{3pt}
\begin{tabular}{p{1.8cm}p{5.2cm}}
\toprule
\multicolumn{2}{l}{\textbf{Irregular Pattern Agent}} \\
\midrule
Role & Third specialist handling noise/anomalies not explained by trend or seasonality \\
Task & Generate concise text description converted to numerical embedding \\
\midrule
\multicolumn{2}{l}{\textit{Analysis priorities (ordered)}} \\
1. Noise & Variance level: high/low/changing \\
2. Anomalies & Outliers, spikes present? \\
3. Autocorr. & Correlated or white noise? \\
4. Difficulty & Correction predictability \\
5. Distribution & Skewness, tail behavior \\
\midrule
Output & JSON: \texttt{\{"irregular\_analysis": "..."\}} \\
Encoding & Text $\rightarrow$ GPT-2 $\rightarrow$ $d_I$-dim embedding \\
\bottomrule
\end{tabular}
\caption{Irregular Agent prompt structure.}
\label{tab:prompt_irregular}
\end{table}

\subsection{Test-Time Adaptation}

Test-time adaptation detects distribution shift between held-out training and test data using STL decomposition statistics. When shift exceeds threshold $\tau$, agents generate adaptation parameters.

\begin{table}[h!]
\centering
\scriptsize
\setlength{\tabcolsep}{3pt}
\begin{tabular}{p{2.2cm}p{4.8cm}}
\toprule
\multicolumn{2}{l}{\textbf{Trend/Seasonal Adaptation}} \\
\midrule
Input & Component shift analysis: validation slope/dominance vs test, shift ratio (\%), direction change \\
\midrule
\multicolumn{2}{l}{\textit{Output (JSON)}} \\
\texttt{shift\_diagnosis} & Description of detected shift \\
\texttt{impact\_assessment} & Expected effect on predictions \\
\texttt{scale\_multiplier} & Multiplicative adjustment to scale \\
\texttt{bias\_shift} & Additive adjustment to bias \\
\texttt{gate\_multiplier} & Multiplicative adjustment to gate \\
\texttt{confidence\_penalty} & Reduction in confidence $\in [0, 0.5]$ \\
\texttt{reasoning} & Explanation of adaptation strategy \\
\bottomrule
\end{tabular}
\caption{Trend/Seasonal adaptation prompt structure.}
\label{tab:prompt_adapt_ts}
\end{table}

\begin{table}[h!]
\centering
\scriptsize
\setlength{\tabcolsep}{3pt}
\begin{tabular}{p{2.2cm}p{4.8cm}}
\toprule
\multicolumn{2}{l}{\textbf{Irregular Adaptation}} \\
\midrule
Input & Original analysis text, statistics comparison (std, mean, spike ratio, autocorr) for validation vs test \\
\midrule
\multicolumn{2}{l}{\textit{Output (JSON)}} \\
\texttt{needs\_adjustment} & Boolean flag \\
\texttt{updated\_text} & Revised irregular analysis \\
\texttt{adapted\_params} & \{scale, bias, gate, confidence\} adjustments \\
\texttt{reasoning} & Explanation \\
\bottomrule
\end{tabular}
\caption{Irregular adaptation prompt structure.}
\label{tab:prompt_adapt_irreg}
\end{table}

\begin{table}[h!]
\centering
\scriptsize
\setlength{\tabcolsep}{3pt}
\begin{tabular}{p{2.2cm}p{4.8cm}}
\toprule
\multicolumn{2}{l}{\textbf{Unified Distribution Shift Adaptation}} \\
\midrule
\multicolumn{2}{l}{\textit{Input: STL decomposition comparison}} \\
Trend shift & Validation vs test slope, shift ratio, significance \\
Seasonal shift & Validation vs test dominance, shift ratio, significance \\
Raw statistics & Mean, std comparison, overall shift score \\
\midrule
\multicolumn{2}{l}{\textit{Output (JSON)}} \\
Base params & shift\_diagnosis, impact\_assessment, scale\_multiplier, bias\_shift, gate\_multiplier, confidence\_penalty \\
Component adj. & trend\_scale\_adjust, seasonal\_scale\_adjust \\
\texttt{reasoning} & Adaptation rationale \\
\midrule
\multicolumn{2}{l}{\textit{Guidance}} \\
Trend shift & Use bias\_shift for systematic change; trend\_scale\_adjust for magnitude \\
Seasonal shift & Use seasonal\_scale\_adjust for magnitude \\
\bottomrule
\end{tabular}
\caption{Unified adaptation prompt structure.}
\label{tab:prompt_adapt_unified}
\end{table}

\end{document}